\documentclass[10pt,twocolumn,dvipsnames]{article}

\usepackage{iccv}
\usepackage{times}
\usepackage{epsfig}
\usepackage{graphicx}
\usepackage{amsmath}
\usepackage{amssymb}
\usepackage{float}
\usepackage{stfloats}
\usepackage{xcolor}
\definecolor{mygray}{RGB}{122, 124, 127}
\definecolor{mygreen}{RGB}{016, 070, 128}
\definecolor{myyellow}{RGB}{250,175,66}
\definecolor{myred}{RGB}{194,69,83}
\usepackage{balance}
\usepackage[accsupp]{axessibility}  %
\usepackage{gensymb}

\usepackage{booktabs}

\usepackage[pagebackref=true,breaklinks=true,colorlinks,bookmarks=false,hypertexnames=false]{hyperref}

\iccvfinalcopy 

\def\iccvPaperID{12755}

\ificcvfinal\fi

\begin{document}

\title{Rethinking pose estimation in crowds: overcoming the detection information bottleneck and ambiguity}

\author{
Mu Zhou\thanks{\textit{Authors contributed equally to this work.}}
~~~~~~~~
Lucas Stoffl\footnote[1] ~~~~~~~~
Mackenzie Weygandt Mathis ~~~~~~~~
Alexander Mathis\\
École Polytechnique Fédérale de Lausanne (EPFL)\\
{\tt\small \ alexander.mathis@epfl.ch}
}

\maketitle

\ificcvfinal\thispagestyle{empty}\fi


\begin{abstract}
Frequent interactions between individuals are a fundamental challenge for pose estimation algorithms. Current pipelines either use an object detector together with a pose estimator (top-down approach), or localize all body parts first and then link them to predict the pose of individuals (bottom-up). Yet, when individuals closely interact, top-down methods are ill-defined due to overlapping individuals, and bottom-up methods often falsely infer connections to distant bodyparts. Thus, we propose a novel pipeline called bottom-up conditioned top-down pose estimation (BUCTD) that combines the strengths of bottom-up and top-down methods. Specifically, we propose to use a bottom-up model as the detector, which in addition to an estimated bounding box provides a pose proposal that is fed as condition to an attention-based top-down model. We demonstrate the performance and efficiency of our approach on animal and human pose estimation benchmarks. On CrowdPose and OCHuman, we outperform previous state-of-the-art models by a significant margin. We achieve 78.5 AP on CrowdPose and 48.5 AP on OCHuman, an improvement of 8.6\% and 7.8\% over the prior art, respectively. Furthermore, we show that our method strongly improves the performance on multi-animal benchmarks involving fish and monkeys. The code is available at \href{https://github.com/amathislab/BUCTD}{https://github.com/amathislab/BUCTD}
\vspace{-10pt}
\end{abstract}

\section{Introduction}
\label{sec:intro}

Imagine somebody hands you an image of a person and asks you ``to annotate the pose". For your exquisite primate visual system this is a trivial task that you can readily achieve. Now imagine somebody hands you another image that contains two people, arm-in-arm. You are likely frustrated and will ask whose pose you should annotate? In response to whose pose you should annotate, your opponent will likely point at the person she has in mind. Based on the pointing, it's again easy to annotate the right pose. Our work proposes a hybrid deep learning framework for pose estimation that is inspired by this interaction.

\begin{figure}
\begin{center}
\includegraphics[width=0.48\textwidth]{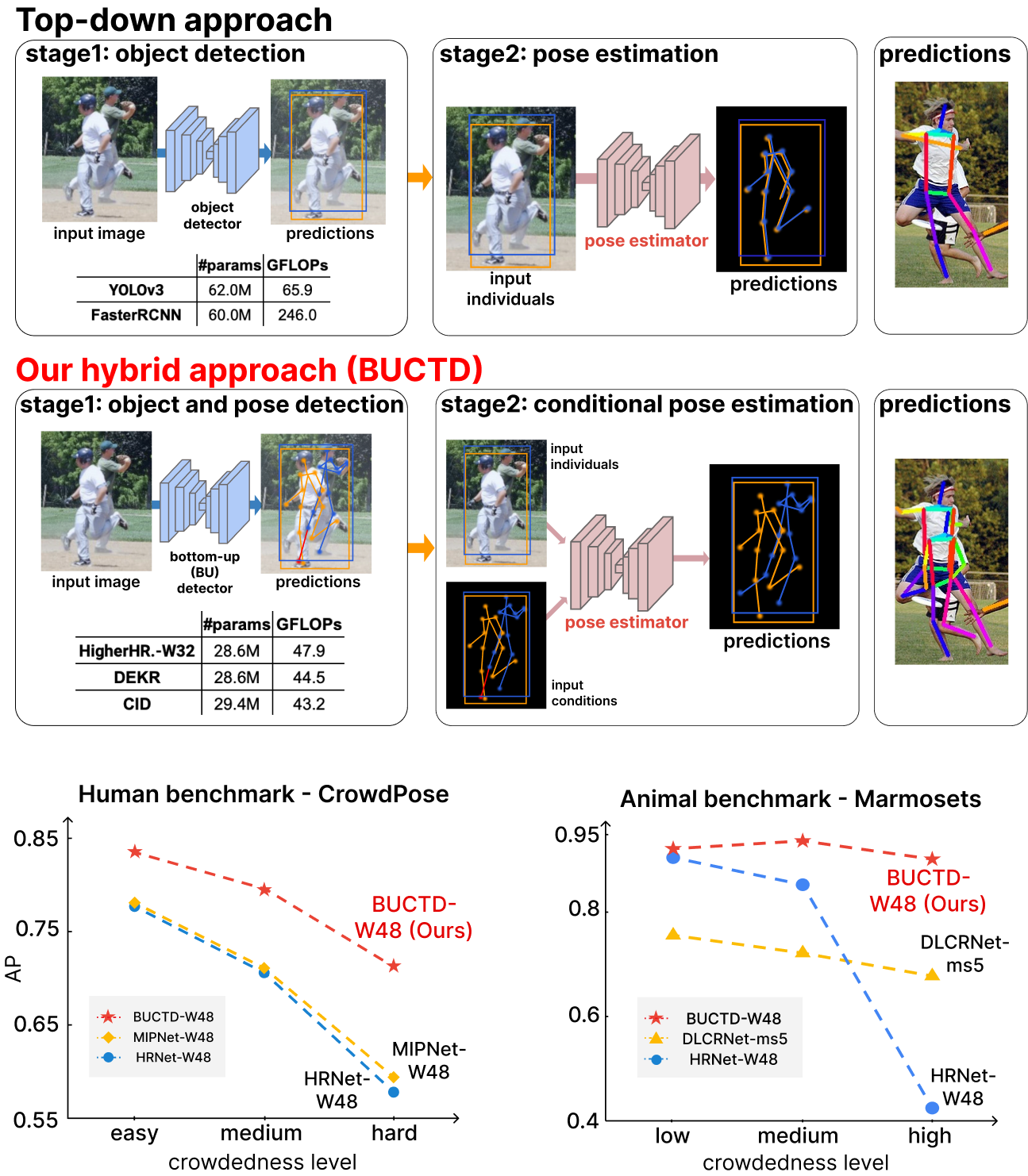}
\end{center} 
\vspace{-5pt}
\caption{\textbf{Overview of our bottom-up conditioned top-down pose estimation (BUCTD) approach and benchmarking results}. BUCTD uses a bottom-up pose model as instance detector, which is computationally cheaper than existing, widely-used object detectors (see inset Table). The pose proposals from the pose detector are used to calculate bounding boxes and to condition our novel, conditional top-down stage. Note that, as in a standard top-down paradigm, only one image crop plus its corresponding conditional pose is presented to the BUCTD. We can substantially boost performance on both human \& animal benchmarks, with especially large gains in crowded scenes (Tables~\ref{table:table_results_crowdpose_sota},~\ref{table:table_results_animals}).}
\label{fig:fig1}
\end{figure}

This simple interaction highlights the ambiguity problem of top-down approaches in crowds. They first localize individuals with a dedicated object detector~\cite{redmon2018yolov3,ren2015faster,yolov5,liu2021Swin} and then perform single-instance pose estimation~\cite{fang2018rmpe,chen2020alphatracker, wang2020deep,li2019crowdpose,yang2021transpose,li2021tokenpose,xu2022vitpose}. In contrast, bottom-up approaches first localize all body parts in the image and then assemble them into poses of each of the individuals simultaneously~\cite{InsafutdinovAPT16,cao2019openpose,newell2017associative,cheng2020higherhrnet,GengSXZW21,lauer2022multi,shi2022end}. Yet, when individuals closely interact, top-down methods are ill-defined as it is unclear which pose should be predicted within a bounding box that contains multiple individuals. Therefore, occluded individuals will often be ignored by top-down methods (Figure ~\ref{fig:fig1}). In contrast, as bottom-up approaches reason over the complete scene they may not have this problem. Bottom-up approaches can localize all individuals, but often struggle to make accurate predictions.

To overcome those limitations we propose a simple yet effective framework called \textbf{B}ottom-\textbf{U}p \textbf{C}onditioned \textbf{T}op-\textbf{D}own pose estimation (BUCTD). Our solution, is inspired by the interaction that we described. Instead of using object detectors, we propose to use bottom-up pose estimation models as detectors. The output poses are used to estimate bounding boxes of the individuals, and also serve as a ``pointing" mechanism, that indicates whose pose should be predicted. To also process the ``pointing" input, we generalize top-down models to conditional-top down (CTD) models, which present the second stage of our BUCTD framework. CTD models take a cropped image together with a pose as input. They are trained to predict the correct pose based on the (potentially) noisy pose provided by the bottom-up methods (Figure~\ref{fig:fig1}). 

Thus, BUCTD overcomes the information bottleneck and ambiguity introduced by standard detectors, while typically having similar or lower inference cost (Figure~\ref{fig:fig1}). We evaluate BUCTD on COCO~\cite{lin2014microsoft}, two crowded human benchmarks, CrowdPose~\cite{li2019crowdpose} and OCHuman~\cite{zhang2019pose2seg}, and three multi-animal benchmarks, namely SchoolingFish, Tri-Mouse and Marmosets~\cite{lauer2022multi}. We achieve SOTA performance and strongly outperform both top-down and bottom-up models in occluded and crowded scenes.

\section{Related Work}
\label{sec:relWork}

\subsection{Multi-instance pose estimation \& benchmarks}

Top-down approaches detect the body parts of each individual by a single-instance pose estimation model~\cite{fang2018rmpe,chen2020alphatracker,wang2020deep} within the detected bounding box generated by the object detector~\cite{ren2015faster, redmon2018yolov3, he2017mask,yolov5,liu2021Swin}. Recently, transformer-based top-down methods such as TransPose~\cite{yang2021transpose}, TokenPose~\cite{li2021tokenpose}, TFPose~\cite{mao2021tfpose} and ViTPose~\cite{xu2022vitpose} have increased the performance. Exemplar bottom-up approaches include OpenPose~\cite{cao2019openpose}, Associative Embedding~\cite{newell2017associative}, ArtTrack~\cite{InsafutdinovAPT16}, HigherHRNet~\cite{cheng2020higherhrnet}, DEKR~\cite{GengSXZW21}, DLCRNet~\cite{lauer2022multi}, CID~\cite{wang2022contextual} and PETR~\cite{shi2022end}. Additionally, ensuring precision in pose estimation has led to the development of pose refinement methods. Works such as PoseRefiner~\cite{fieraru2018learning} and PoseFix~\cite{moon2019posefix} proposed models for refining the predicted pose (of a different model) and can substantially improve the accuracy.

Classic benchmarks for human pose estimation, such as COCO~\cite{lin2014microsoft} and MPII~\cite{andriluka14cvpr}, contain few occlusions~\cite{khirodkar2021multi}, even though this is typical in many real-world scenarios. In recent years new benchmarks with more crowded scenes emerged, most notably CrowdPose~\cite{li2019crowdpose} and OCHuman~\cite{zhang2019pose2seg}. Interestingly, multi-animal pose estimation shares some of the challenges of human benchmarks, but also raises other problems~\cite{mathis2020deep}, such as lack of ``social" distancing amongst animals and highly similar appearances within a given species, such as mice. Therefore, to tackle these challenges we also focused on multi-animal benchmarks comprising mice, monkeys, and groups of fish with heavy overlap~\cite{lauer2022multi}. Our method, BUCTD, achieves SOTA on these benchmarks.

\subsection{Crowded scene pose estimation}

Many recent works~\cite{li2019crowdpose,zhang2019pose2seg,qiu2020peeking,khirodkar2021multi} have focused on occluded scenes in pose estimation. Khirodkar et al.~\cite{khirodkar2021multi} propose a hybrid top-down approach called MIPNet, which allows the model to predict multiple people within a given bounding box. MIPNet reached SOTA performance by providing an integer input to indicate which human with respect to the distance from the center should be predicted. In contrast, we hypothesize that providing a pose cue about which individual should be predicted is advantageous. The CID model~\cite{wang2022contextual} proposes an end-to-end architecture including a CNN backbone and a feature decoupling stage to distinguish between individuals. However, the decoupling is only based on the center prediction of individuals. 
CenterGroup~\cite{braso2021center} uses attention to link person centers to body parts, while PETR~\cite{shi2022end} deploys separate, transformer-based decoders for individuals and keypoints respectively. Ding et al.~\cite{ding20222r} recently proposed another attention-based model that appears to have strong performance on CrowdPose and OCHuman. However, the authors evaluate their model differently than the field, i.e., only based on ground-truth bounding boxes, and hence we do not compare it to other approaches.

\begin{figure*}[ht!]
\begin{center}
\includegraphics[width=\textwidth]{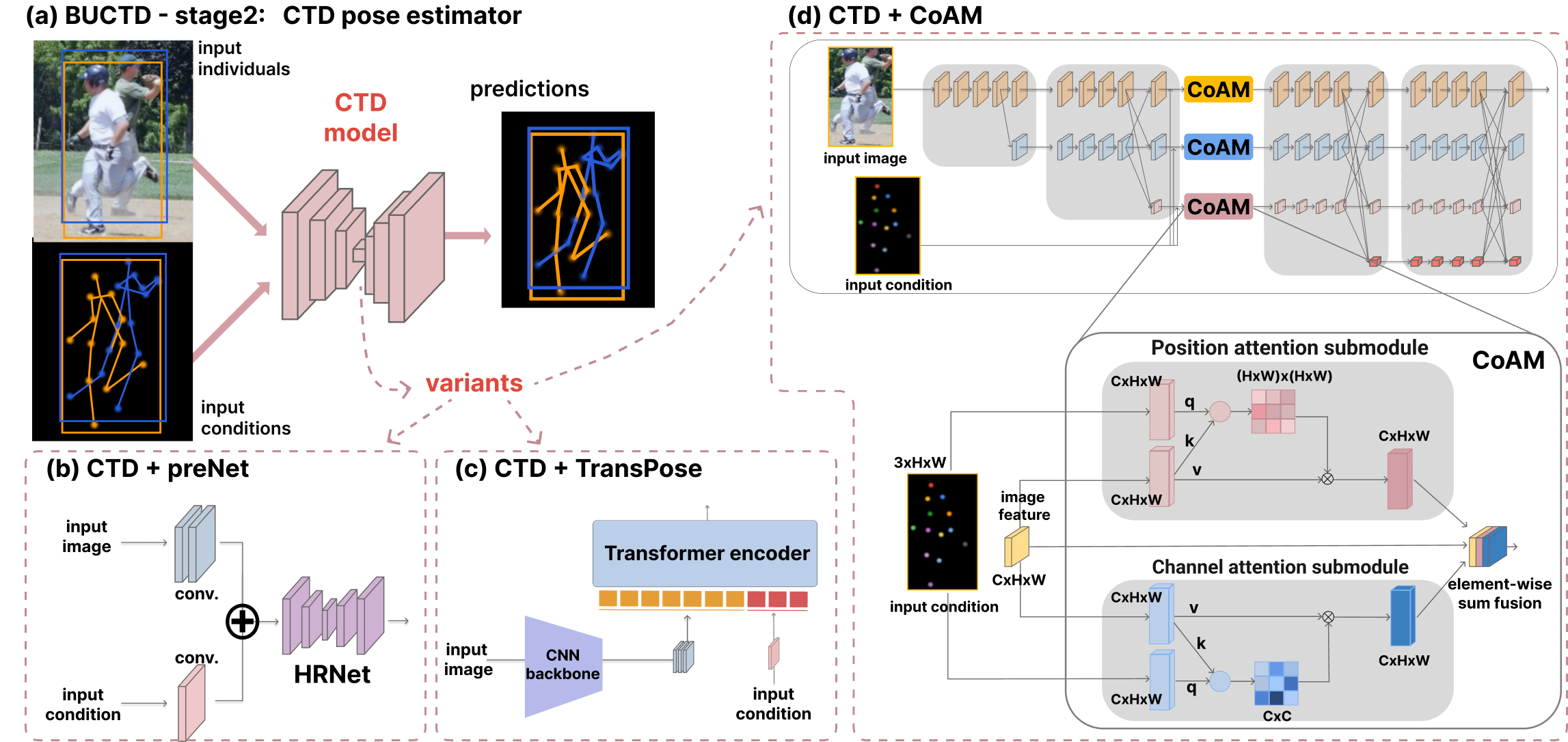}
\end{center} 
\caption{\textbf{Overview of the second stage of our BUCTD approach: conditional top-down (CTD) pose estimators.}
{\bf (a)} CTD receives a conditional pose and a cropped image, whose crop was estimated from the conditioned keypoints. {\bf (b)} CTD with preNet. {\bf (c)} CTD with TransPose. 
{\bf (d)} CTD with Conditional Attention Module (CoAM). {\bf (Top:} Our CTD model simply employs the same multi-resolution stages as the standard HRNet; we insert our CoAM module after stage 2 (one CoAM for every resolution branch).  {\bf Bottom}: CoAM. The output of stage 2 for one branch (image feature) is treated as keys and values for the two attention submodules, while we feed the color-coded condition heatmap as queries into the attention mechanisms. The output of the module is a combination of the spatial attention feature, the channel attention feature and the original image feature.}
\label{fig:BUCTD_model}
\end{figure*}

\subsection{Combining Top-Down and Bottom-Up Models}

Hu and Ramanan~\cite{hu2016bottom} proposed a bidirectional architecture for hierarchical Rectified Gaussian models incorporating top-down feedback with a bottom-up architecture, while Tang et al.~\cite{tang2018deeply} introduced a hierarchical, compositional model, for which the inference process consists of both bottom-up and top-down stages across multiple semantic levels. Cai et al.~\cite{cai2019exploiting} developed a graph-based method for 3D pose estimation by concatenating bottom-up features and top-down features together. Li et al.~\cite{li2019multi} proposed to use bottom-up methods to estimate the joints and leverage the bounding boxes from an object detector to group the joints, while Cheng et al.~\cite{cheng2022dual} shows another, similar way to combine top-down and bottom-up approaches. 

In comparison to previous works, instead of using an object detector, BUCTD leverages bottom-up models as detectors to provide a `pointer' to guide the adapted TD model to pay attention to the correct target individual. For the first stage of BUCTD, we are building on the latest bottom-up methods, such as CID~\cite{wang2022contextual}, and PETR~\cite{shi2022end}. For the second stage, we generalized top-down models to conditional top-down models (CTD). Instead of an index as in MIPNET~\cite{khirodkar2021multi}, we provide a ``pointer'' in the form of a pose predicted from a bottom-up model. Thus, our CTD model gets two inputs: a bounding box and a pose, similar in spirit to PoseFix~\cite{moon2019posefix} or PoseRefiner~\cite{fieraru2018learning}. Akin to PoseFix and PoseRefiner, we also test a convolutional PreNet that provides the pose at the early stage of the top-down model. Moreover, we developed transformer and attention-based models for providing the conditional input. We find that these models achieve better performance and are efficient. 

How is the conditional input provided to the second-stage model (during training)? We either sample it from known error distributions~\cite{ruggero2017benchmarking,moon2019posefix}, or based on the predictions of the BU models. As we will show, both approaches provide strong results.

\section{Methods}
\label{sec:BUCTD}

\begin{figure*}[ht!]
\begin{center}
\includegraphics[width=\textwidth]{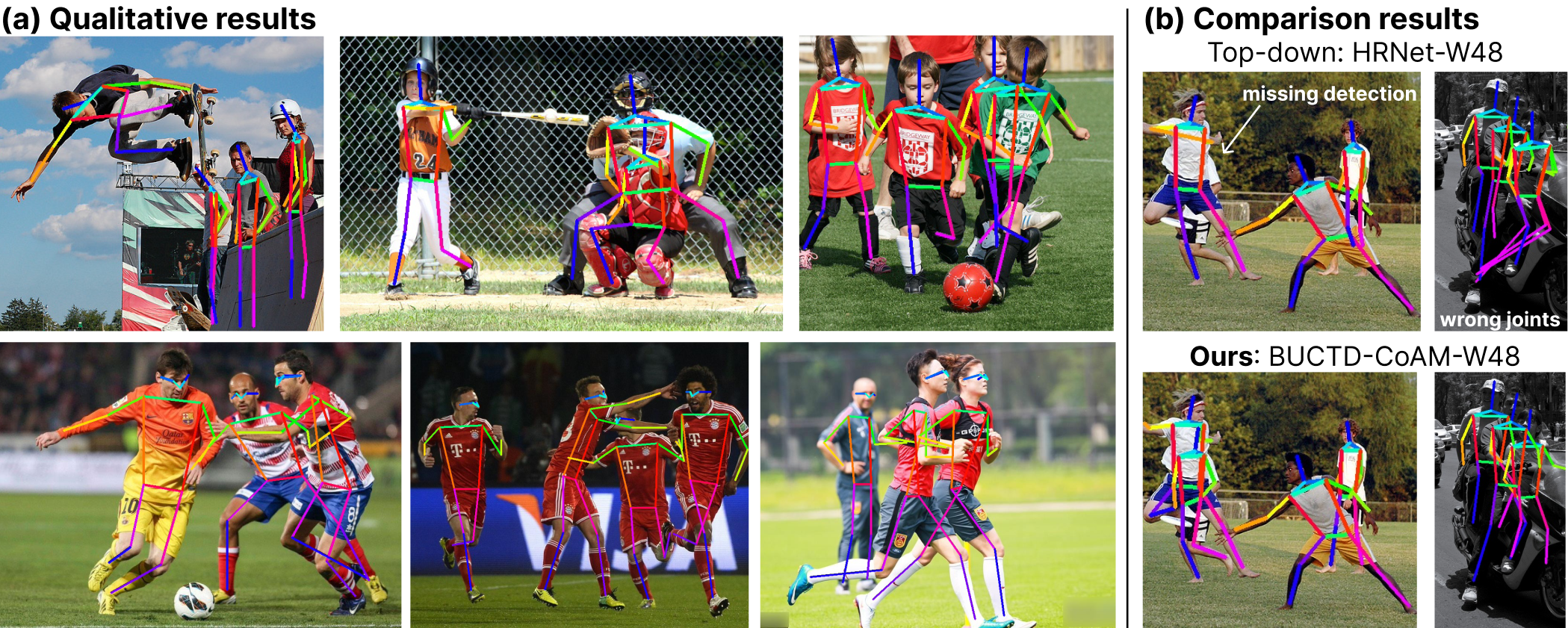}
\end{center} 

\caption{\textbf{CrowdPose and OCHuman, qualitative results.} {\bf (a)} Qualitative results on CrowdPose (top row) and OCHuman (bottom row) with BUCTD-CoAM-W48. {\bf (b)} Top: Predictions from a top-down approach (HRNet-W48) and bottom: predictions from our BUCTD model with CoAM-W48, both on CrowdPose images. Missing detections or wrongly predicted keypoints are noted.}
\label{fig:vis_results}
\end{figure*}

\subsection{The BUCTD framework}
\label{subsec:matching}

Our BUCTD model is a two-stage model trained to predict the pose from the cropped input image and the bottom-up input (Figure~\ref{fig:BUCTD_model}). The intuition is that the BUCTD model can use the image as well as the pose input to ``know'' which individual to predict. This overcomes the inherent ambiguity for top-down approaches in crowded scenes. Importantly, our training scheme differs from classic top-down approaches, since we train on cropped images generated from bottom-up pose predictions, in contrast to using the ground-truth bounding boxes. This induces additional augmentation in the training.

\textbf{Stage1: Bottom-Up detector (BU).} Firstly, to detect individuals, we used bottom-up pose estimation models on the target training dataset and get the predictions. Classic top-down methods use a generic object detector to get the bounding boxes for individuals, however, bounding boxes create an information bottleneck between the detector and the pose estimation model. Furthermore, in real-world applications, training an object detector often has a higher computational cost compared to training bottom-up pose models (Suppl.\ Materials Section~\ref{sec:compcosts}). Therefore, we propose to use a bottom-up model as the detector, which will provide both a predicted pose as guidance, and the corresponding bounding box. Of course, also recent single-stage models~\cite{shi2022end,wang2022contextual} can (and will) be used.

\textbf{Stage2: Conditional Top-Down (CTD).} Secondly, we train CTD models with the conditional pose input and the corresponding bounding box. During training, we experiment with two different sampling strategies for the conditions. 1) {\bf Empirical sampling:} We sample the conditions from the predictions of BU models, where we match those predictions to the ground truth pose by using Object Keypoint Similarity (OKS) on human benchmarks and bounding box IoU on animal benchmark. 2) {\bf Generative sampling:} Instead of taking actual predictions, we also synthetically sample input poses based on estimated pose estimation errors~\cite{ruggero2017benchmarking}. This approach is similar to PoseFix~\cite{moon2019posefix}.

\subsection{CTD model architectures}
\label{subsec:models}

The second stage of our BUCTD approach consists of the conditional top-down (CTD) pose estimation model, that provides a generic solution to tackle multi-instance pose estimation in crowded scenes. In this section, we describe the flexibility of the BUCTD approach by adopting different TD architectures (Figure~\ref{fig:BUCTD_model}a-d). In all cases, we convert the conditional input from the predicted keypoints into a 3-channel heatmap by using a Gaussian distribution with a standard deviation $\sigma$. To keep the semantic information, we provide each keypoint with a certain RGB value. We validate these design choices in the Suppl.\ Materials Section~\ref{subsec:ValidationHM}. 

\subsubsection{CTD with preNet}

Given an input pair comprising an input image and input condition, we feed the pair in parallel to the preNet which contains two different convolutional layers, to extract features from both inputs (i.e., two 7x7 conv. layers for input image and one 7x7 conv. layer for the conditional input). Then, we fuse the image feature and the condition feature together and feed it into a HRNet~\cite{wang2020deep}. Thus, CTD-preNet is a simple architecture leveraging the CTD approach (Figure~\ref{fig:BUCTD_model}b), which is similar to PoseFix's design~\cite{moon2019posefix}.

\subsubsection{CTD with TransPose}

We modified TransPose~\cite{yang2021transpose} to leverage its powerful transformer architecture. TransPose consists of a CNN backbone whose output features are transformed to $d \times H \times W$ by a $1\times1$ convolution. These d-dimensional image feature maps are then flattened into a sequence $X \in \mathbb{R}^{N \times d}$, with $N = H \times W$, which is then fed to a standard transformer encoder. For CTD+TransPose, we provide conditions as ``side-information'' by concatenating condition-specific tokens to this transformer input sequence (Figure~\ref{fig:BUCTD_model}c). Specifically, we transform the conditional heatmap first by a $1\times1$ convolution to expand the number of channels to $c$ and, after flattening, obtain a condition sequence $Y \in \mathbb{R}^{N \times c}$. Here, we arbitrarily chose $c=16$. The final input sequence to the transformer encoder is the concatenation of $X$ and $Y$, i.e., $X \oplus Y$. The attention layers enable capturing long-range relationships between the conditional input and the predicted keypoints.

\subsubsection{CTD with a Conditional Attention Module}

To learn a better representation and leverage the information from the conditional input, we propose an architecture comprised of a Conditional Attention Module (CoAM) and HRNet~\cite{wang2020deep}, where we feed the input image and the corresponding pose condition into HRNet and the CoAM in parallel (Figure~\ref{fig:BUCTD_model}d). The CoAM can be inserted after any HRNet-stage and its output is fused with the features of the corresponding stage. Unless otherwise indicated the CoAM input was given to HRNet-stage 2 (see ablations). Therefore, the conditional input (1) provides a clue to which individual in the crop the CTD model should focus on, (2) improves the input pose obtained from the bottom-up model.

CoAM is inspired by~\cite{fu2019dual} and contains spatial and channel attention sub-modules. It is designed to learn associations between features and ``conditions" with an attention-like mechanism. Finally, it aggregates and fuses the features from both sub-modules by performing an element-wise sum. The resulting feature map is added back to the module's input feature map, hence combining the features extracted from the HRNet with the long-range contextual information. The CoAM module treats the conditions as queries and the feature maps as keys and values for calculating the attention scores (Figure~\ref{fig:BUCTD_model}d). Details on the Position and Channel Attention Modules are provided in the Suppl.\ Materials Section~\ref{subsec:CoAMdetails}.

\subsection{Implementation \& training details}

\textbf{BUCTD with preNet.} We equipped HRNet-W32 or W48~\cite{wang2020deep} with a preNet to train BUCTD-preNet-W32 and W48, respectively, and report the efficiency on the animal benchmarks~\cite{lauer2022multi}, CrowdPose~\cite{li2019crowdpose} and COCO~\cite{lin2014microsoft}.

\textbf{BUCTD with TransPose.}
We trained BUCTD-TP-H-A6 model (based on the TransPose-H-A6 architecture~\cite{yang2021transpose}) on CrowdPose to show that the conditional top-down approach can be successfully integrated into transformers. We ran this architecture on CrowdPose~\cite{li2019crowdpose} and COCO~\cite{lin2014microsoft}.

 \textbf{BUCTD with CoAM.}
We trained HRNet-W32 and HRNet-W48~\cite{wang2020deep} with CoAM (BUCTD-CoAM-W32 and BUCTD-CoAM-W48) on all benchmarks~\cite{lauer2022multi,li2019crowdpose,lin2014microsoft,zhang2019pose2seg}. 
(For additional details see Suppl.\ Material Section~\ref{sec:implementation_details}.)

\textbf{Training Details:} 
To obtain the conditional inputs we trained bottom-up pose estimation models: specifically DLCRNet~\cite{lauer2022multi} for animal pose, and HigherHRNet~\cite{cheng2020higherhrnet} for CrowdPose and OCHuman, and saved (pose) predictions from different model checkpoints (animals: up to 8-12 checkpoints, human: 15 checkpoints).

During training and inference, we added a fixed margin (25 pixels) in height and width to the predicted bounding box, for animal datasets. To keep the aspect ratio and avoid distortion of the animal's body, we resized and padded the predicted bounding box to 256$\times$256. For the human data, we used a margin of 5 pixels and extend each detection box to a fixed aspect ratio (256$\times$192 or 384$\times$288).
We followed the same training scheme (batch size, learning rate, weight initialization, augmentation scheme, loss function) as in~\cite{wang2020deep, khirodkar2021multi} (see Suppl.\ Materials Section~\ref{sec:implementation_details}).
As for the experiments using generative sampling, we use the same error distribution as in PoseFix~\cite{ruggero2017benchmarking,moon2019posefix} on human benchmarks, while for animal benchmarks we adapt it (Suppl.\ Materials Section~\ref{subsec:generativesampling}).

\section{Experiments}
\label{sec:experiments}

To evaluate BUCTD, we performed comprehensive experiments on several benchmarks. We tested our approach on the most important benchmarks for crowded scenes (CrowdPose~\cite{li2019crowdpose} and Occluded Human (OCHuman)~\cite{zhang2019pose2seg}), as well as on three multi-animal pose estimation benchmarks~\cite{lauer2022multi}, and COCO~\cite{lin2014microsoft}.
We also carried out several ablations to test the design choices.

\subsection{CrowdPose Benchmark}

\textbf{Dataset:} The CrowdPose dataset~\cite{li2019crowdpose} contains $12K$ labeled images in the $trainval$ set with $43.4K$ labeled people (each with 14 keypoints), and $8K$ images in the $test$ set with $29K$ labeled people. Following other studies~\cite{GengSXZW21, shi2022end, wang2022contextual}, we used $trainval$ for training, and $test$ for evaluation. We report standard metrics AP, AP${_{easy}}$, AP${_{med}}$ and AP${_{hard}}$ as defined in~\cite{li2019crowdpose}. We compared our method, that derives bounding boxes from a bottom-up model (see Methods), with baselines that used bounding boxes obtained by a Faster R-CNN detector~\cite{ren2015faster}.

\textbf{Results:} 
First, we trained the different CTD models on CrowdPose with BU predictions from HigherHRNet-W32 and empirical sampling. We found that all variants can boost the performance of HigherHRNet by up to 7 AP (Table~\ref{table:table_results_crowdpose_errcorr}). We evaluated the test-performance (without re-training) of CTD models when they were provided with inputs from recently described SOTA bottom-up or single-stage models (CID-W32, DEKR and PETR, respectively). We found that the CTD models could generalize to other bottom-up pose estimation model inputs (Table~\ref{table:table_results_crowdpose_errcorr}). Using the CoAM module provided the best results, and outperformed the CTD-preNet models, which is similar to PoseFix~\cite{moon2019posefix}. This highlights that CTD models are good pose refiners. Next we checked the performance of the full BUCTD pipeline. 

We then compared BUCTD to bottom-up, single-stage, and top-down methods. Overall, BUCTD achieved SOTA performance on CrowdPose (Table~\ref{table:table_results_crowdpose_sota}, Figures~\ref{fig:vis_results},~\ref{fig:vis_results_2} for failure cases~\ref{fig:failures}). As we already showed, BUCTD can improve the performance over BU models, at the cost of additional computation (due to inference with CTD models). BUCTD also outperforms top-down and hybrid methods, while having comparable computational costs. Training and performing inference are comparable for object detectors and BU models, as BU models often have similar or fewer parameters and GFLOPs (Suppl.\ Materials Section~\ref{sec:compcosts}). 

Strikingly, we improved upon MIPNet-W48~\cite{khirodkar2021multi} by up to 8.5 AP.
To achieve this, BUCTD-CoAM-W48 was trained with bottom-up predictions from HigherHRNet-W32~\cite{cheng2020higherhrnet} (which alone performs relatively poorly) and generative sampling, as well as inputs from PETR~\cite{shi2022end}. We also compared to PoseFix~\cite{moon2019posefix}, which did not evaluate on CrowdPose~\cite{li2019crowdpose}. Thus, we implemented PoseFix by utilizing generative sampling and our BUCTD-preNet architecture, which is similar to the original PoseFix~\cite{moon2019posefix}, but with a more powerful backbone. This PoseFix-HRNet-W48 achieves 76.8 AP, while BUCTD (with CoAM) achieves 78.5 AP. Interestingly, for BUCTD, we also achieved the best results with generative sampling. Collectively, this suggests that our hybrid approach combines the strengths of both bottom-up and top-down methods, and can outperform refinement methods.

Moreover, to gain additional insights, we computed the precision and recall for the BUCTD approach for different BU models on CrowdPose. We compare our model to the previous SOTA on CrowdPose: MIPNet~\cite{khirodkar2021multi}. Importantly, BUCTD has both higher recall and precision than MIPNet for all models (Suppl.\ Materials Section~\ref{subsec:precisionrecall}).

\setlength{\tabcolsep}{3.pt}
\begin{table}[t!]
\centering
\footnotesize
\begin{tabular}{l|c|ccc}
\hline
Method & AP & AP$_{easy}$ & AP$_{med}$ & AP$_{hard}$ \\
\hline
HigherHRNet-W32~\cite{cheng2020higherhrnet} & 65.7 & 73.2 & 66.1 & 57.9 \\
CTD-preNet-W32$\dagger$  & 69.5 (+3.8) & 76.2 & 69.9 & 62.5 (+4.6)\\
CTD-TP-H-A6$\dagger$ & 70.7 (+5.0) & 77.9 & 71.1 & 63.0 (+5.1) \\
CTD-CoAM-W32$\dagger$ & 71.4 (+5.7) & 78.0 & 71.8 & 64.5 (+6.6) \\
CTD-CoAM-W48$\star$ & 72.9 (+7.2) & 79.2 & 73.4 & 66.1 (+8.2) \\
\hline
DEKR~\cite{GengSXZW21} & 68.0 & 76.6 & 68.8 & 58.4 \\
CTD-preNet-W32$\dagger$ & 69.7 (+1.7) & 77.7 & 70.6 & 60.5 (+2.1) \\
CTD-TP-H-A6$\dagger$ & 71.0 (+3.0) & 79.1 & 71.9 & 61.7 (+3.3) \\
CTD-CoAM-W32$\dagger$ & 71.1 (+3.1) & 78.8 & 71.9 & 61.8 (+3.4) \\
CTD-CoAM-W48$\star$ & 72.0 (+4.0) & 79.5 & 72.8 & 63.0 (+4.6) \\
\hline
CID-W32~\cite{wang2022contextual} & 71.3 & 77.4 & 72.1 & 63.9 \\
CTD-preNet-W32$\dagger$ & 72.8 (+1.5) & 79.0 & 73.4 & 65.7 (+1.8)\\
CTD-TP-H-A6$\dagger$ & 73.7 (+2.4) & 80.1 & 74.5 & 66.2 (+2.3) \\
CTD-CoAM-W32$\dagger$ & 74.2 (+2.9) & 80.2 & 74.9 & 67.1 (+3.2) \\
CTD-CoAM-W48$\star$ & 75.3 (+4.0) & 81.1 & 75.9 & 68.4 (+4.5) \\
\hline
PETR~\cite{shi2022end} & 72.0 & 78.0 & 72.5 & 65.4 \\
CTD-preNet-W32$\dagger$ & 74.6 (+2.6) & 80.9 & 75.1 & 67.7 (+2.3)\\
CTD-TP-H-A6$\dagger$ & 75.6 (+3.6) & 82.2 & 76.1 & 68.6 (+3.2) \\
CTD-CoAM-W32$\dagger$ & 75.9 (+3.9) & 81.9 & 76.3 & 69.1 (+3.7) \\
CTD-CoAM-W48$\star$ & 76.7 (+4.7) & 82.6 & 77.2 & 70.4 (+5.0) \\
\hline
\end{tabular}
\vspace{3pt}
\caption{\textbf{CTD boosts CrowdPose results using conditional inputs from different bottom-up and single-stage models} (on $test$ set). $\dagger$ and $\star$ denotes input resolution of 256x192 and 384x288 respectively. All CTD models are trained with HigherHRNet-W32 conditional input.}
\label{table:table_results_crowdpose_errcorr}
\end{table}

\setlength{\tabcolsep}{1pt}
\begin{table}[!ht]
\centering
\footnotesize
\begin{tabular}{l|c|c|ccc}
\hline
Method & Input size & AP & AP$_{easy}$ & AP$_{med}$ & AP$_{hard}$ \\
\hline
\multicolumn{2}{l}{Bottom-Up methods}\\
\hline
HRNet-W48~\cite{wang2020deep} & 640 & 67.3 & 74.6 & 68.1 & 58.7 \\
HigherHRNet-W48~\cite{cheng2020higherhrnet} & 640 & 67.6 & 75.8 & 68.1 & 58.9 \\
CenterGroup~\cite{braso2021center} & 640 & 69.4 & 76.6 & 70.0 & 61.5 \\
BAPose~\cite{artacho2021bapose} & 512 & 72.2 & 79.9 & 73.4 & 61.3 \\
\hline
\multicolumn{2}{l}{Single-stage methods}\\
\hline
DEKR~\cite{GengSXZW21} & 640 & 68.0 & 76.6 & 68.8 & 58.4 \\
PETR~\cite{shi2022end} & 800 & 72.0 & 78.0 & 72.5 & 65.4 \\
CID-W32~\cite{wang2022contextual} & 512 & 71.3 & 77.4 & 72.1 & 63.9 \\
CID-W48~\cite{wang2022contextual} & 640 & 72.3 & 78.7 & 73.0 & 64.8 \\
\hline
\multicolumn{2}{l}{Top-Down methods}\\
\hline
AlphaPose~\cite{fang2018rmpe} & - & 61.0 & 71.2 & 61.4 & 51.1 \\
JC-SPPE~\cite{li2019crowdpose} & 320$\times$256 & 66.0 & 75.5 & 66.3 & 57.4 \\
HRNet-W48~\cite{wang2020deep} & 384$\times$288 & 69.3 & 77.7 & 70.6 & 57.8 \\
\hline
\multicolumn{2}{l}{Hybrid approaches}\\
\hline
MIPNet-W48~\cite{khirodkar2021multi} & 384$\times$288 & 70.0 & 78.1 & 71.1 & 59.4 \\
PoseFix (HRNet-preNet-W48)++$\sigma$ & 384$\times$288 & 76.8 &  82.3 & 77.4 & 70.2   \\
BUCTD-CoAM-W48 (Ours) & 384$\times$288 & 72.9 & 79.2 & 73.4 & 66.1 \\
BUCTD-CoAM-W48+ (Ours) & 384$\times$288 & 75.3 & 81.1 & 75.9 & 68.4 \\
BUCTD-CoAM-W48++ (Ours) & 384$\times$288 & 76.7 & 82.6 & 77.2 & 70.4 \\
\textbf{BUCTD-CoAM-W48++$\sigma$ (Ours)} & 384$\times$288 & {\bf 78.5} & {\bf 83.9} & {\bf 79.0} & {\bf 72.3} \\
\hline
\end{tabular}
\vspace{2pt}
\caption{\textbf{BUCTD improved performance on CrowdPose} $test$ set. For empirical sampling, BUCTD models are trained with HigherHRNet-W32 conditions.  
+ denotes cond. input from CID-W32, ++ denotes cond. input from PETR, and $\sigma$ denotes generative sampling.}
\label{table:table_results_crowdpose_sota}
\end{table}

\textbf{Ablation Results:} 
Some BU models provide more predictions than detectors. To fairly compare, we also provided exactly the same number of detections from the bottom-up models as provided by the object detector. Despite this artificial constraint, the performance of BUCTD was still significantly higher than the one of MIPNet~\cite{khirodkar2021multi}, indicating that the performance gains are not simply coming from a higher number of provided detections (Suppl.\ Materials Section~\ref{subsec:ablation-detections}).

To validate our design choices we performed ablation studies on the CTD-CoAM-W32 model.
We experimented with the position of CoAM and found that conditional inputs are best provided at earlier stages in the network (Table~\ref{table:table_results_crowdpose_ablations_attention}). We further wanted to validate the impact of different conditional inputs during training. In comparison to the 15 checkpoints we obtained from training the bottom-up model (HigherHRNet-W32), we trained CTD models with conditions from different numbers and types of conditions. Generally speaking, CTD models can better learn to predict poses when they are trained with diverse conditional poses for empirical sampling. Accordingly, generative sampling can further improve the performance (Table \ref{table:table_results_crowdpose_ablations_conditions}).

\setlength{\tabcolsep}{3.pt}
\begin{table}[ht]
\centering
\footnotesize
\begin{tabular}{l|c|c|ccc}
\hline
Method & stage & AP & AP$_{easy}$& AP$_{med}$ & AP$_{hard}$ \\
\hline
BUCTD-CoAM-W32 & stage 1 & {\bf 71.2} & {\bf 77.8} & \textbf{71.6} & 64.2 \\
BUCTD-CoAM-W32 & stage 2 & {\bf 71.2} & 77.7 & \textbf{71.6} & {\bf 64.3} \\
BUCTD-CoAM-W32 &  stage 3 &70.5 & 77.4 & 70.9 & 63.3 \\
BUCTD-CoAM-W32 & stage 4 & 68.3 & 76.8 & 69.2 & 58.7 \\
\hline
\end{tabular}
\vspace{2pt}
\caption{\textbf{Ablating the position of CoAM on CrowdPose $test$}. Feeding the conditions through CoAM in earlier stages of the HRNet architecture increased performance.}
\label{table:table_results_crowdpose_ablations_attention}
\end{table}

\begin{table*}[ht]
\centering
\footnotesize
\begin{tabular}{c|c|c|c|c|cccc|cccc}
\hline
GT & \#checkpts & \#checkpts & \#checkpts & \#checkpts & AP & AP$_{easy}$ & AP$_{med}$ & AP$_{hard}$ & AP & AP$_{easy}$ & AP$_{med}$ & AP$_{hard}$ \\[1pt]
 & HrHRNet & DEKR & CID & PETR & \multicolumn{4}{c|}{Tested on HrHRNet conditions} & \multicolumn{4}{c}{Tested on PETR conditions} \\[1pt]
\hline
- & 1 (best) & - & - & - & 68.8 & 75.8 & 69.3 & 61.5 & 73.9 & 80.4 & 74.4 & 67.1 \\[2pt]
- & 1 (test-gap) & - & - & - & 69.8 & 76.6 & 70.3 & 62.8 & 74.7 & 80.9 & 75.1 & 67.9 \\[2pt]
- & 1 & 1 & 1 & 1 & 68.9 & 76.0 & 69.4 & 61.5 & 74.4 & 80.7 & 74.8 & 67.6 \\[2pt]
- & 15 & - & - & - & 71.4 & 78.0 & 71.8 & 64.5 & 75.9 & 81.9 & 76.3 & 69.1 \\[2pt]
1 & 15 & 1 & 1 & 1 & 70.9 & 77.7 & 71.5 & 63.9 & 75.5 & 81.7 & 75.9 & 68.9  \\[2pt]
+gen. & - & - & - & - & 72.3 & 78.8 & 72.8 & 65.1 & 76.9 & 83.0 & 77.4 & 70.4 \\
\hline
\end{tabular}
\vspace{2pt}
\caption{\textbf{Additional ablation studies for the number and type of different checkpoints during training on CrowdPose.} For empirical sampling, higher diversity (predictions sampled from different checkpoints of the bottom-up model) leads to better performance. We also found that generative sampling works well. ``Best'' denotes the best performing checkpoint, while ``test-gap'' denotes the one for which training performance is closest to the final testing performance. All models are based on a HRNet-W32, are trained on input resolutions of 256x192 and tested with flipping.}
\label{table:table_results_crowdpose_ablations_conditions}
\end{table*}

\subsection{OCHuman Benchmark}
\vspace{-4pt}
\textbf{Dataset:} The OCHuman dataset~\cite{zhang2019pose2seg} is the most challenging dataset for crowded multi-person pose estimation with an average of $0.67$ MaxIoU (Intersection over Union between bounding boxes) for each person. It contains $4,731$ images with $8,110$ persons in total. For a fair comparison, we report the results in the same way as illustrated in ~\cite{zhang2019pose2seg,wang2022contextual}, namely we train our models on the COCO $train$ set and evaluate on OCHuman $test$ set.

\textbf{Results:}
Previous SOTA results on OCHuman were reported by the single-stage model CID~\cite{wang2022contextual}. We reach new SOTA performance, with gains up to 3.5 AP. Notably even BUCTD with a smaller HRNet outperforms MIPNet. Naturally, the BUCTD pipeline improves over plain bottom-up methods, which historically do well in crowded scenes, and over the recently introduced strong single-stage method CID~\cite{wang2022contextual}. Qualitative results can be seen in Figure~\ref{fig:vis_results}.

\begin{table}[ht]
\centering
\begin{tabular}{l|cc}
\hline
Method & AP val & AP test \\
\hline
HGG~\cite{jin2020differentiable} & 35.6 & 34.8 \\
HigherHRNet-W32~\cite{cheng2020higherhrnet} & 40.0 & 39.4 \\ 
LOGO-CAP-W48~\cite{xue2022learning} & 41.2 & 40.4 \\  
\hline
DEKR~\cite{GengSXZW21} & 37.9 & 36.5 \\
CID-W32~\cite{wang2022contextual} & 45.7 & 44.6 \\
CID-W48~\cite{wang2022contextual} & 46.1 & 45.0 \\
\hline
AlphaPose+~\cite{qiu2020peeking} & - & 27.5 \\
HRNet-W48$\star$~\cite{wang2020deep} & 37.8 & 37.2 \\
MIPNet-W48$\star$~\cite{khirodkar2021multi} & 42.0 & 42.5 \\
\hline
BUCTD-CoAM-W32 (HrHRNet-W32)$\dagger$ & 44.1 & 43.5 \\
\hline
BUCTD-CoAM-W32 (CID-W32)$\dagger$ & 47.3 & 46.3 \\
BUCTD-CoAM-W48 (CID-W32)$\sigma$$\dagger$ &  48.3 & 47.4 \\
BUCTD-CoAM-W48 (CID-W32)$\sigma$$\dagger$ 2x & 48.8 & 48.3 \\
BUCTD-CoAM-W48 (CID-W32)$\sigma$$\dagger$ 3x & {\bf 49.0} & {\bf 48.5} \\
\hline
\end{tabular}
\vspace{2pt}
\caption{\textbf{BUCTD improved performance on OCHuman.} Comparison with state-of-the-art methods on the OCHuman val and test set after training on COCO train. $\dagger$ and $\star$ denotes input resolution of 256x192 and 384x288, respectively. Model in brackets denotes where conditions are coming from during inference. 2x and 3x marks iterative refinement by feeding back model predictions as new conditions. 
For empirical sampling, BUCTD models are trained with HigherHRNet-W32 conditions. $\sigma$ denotes generative sampling.}
\label{table:table_results_sota_ochuman}
\end{table}

\subsection{COCO Benchmark}

\textbf{Dataset:} The COCO~\cite{lin2014microsoft} dataset contains 57K images with 150K persons in the $train$ set, 5K images with 6.3K persons in the $val$ set and 20K images in the test-dev set. We used $train$ for training and $val$ for validation. We compared our method with several bottom-up models and top-down methods, but note that it has few overlapping people~\cite{khirodkar2021multi} compared to CrowdPose~\cite{li2019crowdpose} and OCHuman~\cite{zhang2019pose2seg}. 

\textbf{Results:} BUCTD can refine the performance of single-stage, BU and TD (HRNet-W48) methods. This suggests that our method, designed for the challenges of crowdedness, can also fare well even with few overlapping people (Table~\ref{table:table_results_coco}).

\begin{table}[hbp!]
\centering
\begin{tabular}{l|ccc}
\hline
Method  & AP & AP$_{M}$ & AP$_{L}$ \\
\hline
DEKR~\cite{GengSXZW21}  & 71.0  & 66.7  & 78.5   \\
CID-W32~\cite{wang2022contextual} &  69.8  &  64.0  &  78.9 \\
PETR~\cite{shi2022end} & 73.1  & 67.2 & 81.7\\
HRNet-W48~\cite{wang2020deep} & 76.3  & 72.3 & 83.4  \\
MIPNet-W48~\cite{khirodkar2021multi} & 76.3  &  72.3 & 83.4  \\
PoseFix~\cite{moon2019posefix} (best original)$\sigma$ &  73.6 & 70.3 & 79.8 \\
PoseFix (our HRNet-preNet-W48)$\sigma$ &  77.3 & 73.5 & 84.4 \\
\hline
BUCTD-CoAM-W48 (DEKR)$\sigma$  &  74.8 & 71.1 &  81.1 \\
BUCTD-CoAM-W48 (CID-W32)$\sigma$  & 74.8 & 71.1 &  81.1 \\
BUCTD-CoAM-W48 (PETR)$\sigma$  &  77.1 &  73.3 & 83.4 \\
\bf BUCTD-preNet-W48 (PETR)$\sigma$ & \bf 77.8  & \bf 74.2  & \bf 83.7 \\
BUCTD-TP-H-A6 (PETR)$\sigma$ & 76.0  & 72.2 & 82.3 \\
BUCTD-CoAM-W48 (HRNet-W48)$\sigma$  & 76.5  & 72.7 &  83.2 \\
\hline
\end{tabular}
\vspace{2pt}
\caption{\textbf{Results on COCO val set}. BUCTD can be used to refine poses from other models also on COCO.
Models trained with generative sampling are denoted by $\sigma$.}
\label{table:table_results_coco}
\end{table}

\subsection{Multi-Animal Benchmarks}

\textbf{Datasets:} To further assess the performance of BUCTD, we evaluated it on multi-animal benchmarks by Lauer et al. called SchoolingFish, Marmosets, and Tri-Mouse~\cite{lauer2022multi}. 

\begin{figure}[ht]
\begin{center}
\setlength{\abovecaptionskip}{0.1cm}
\includegraphics[width=0.47\textwidth]{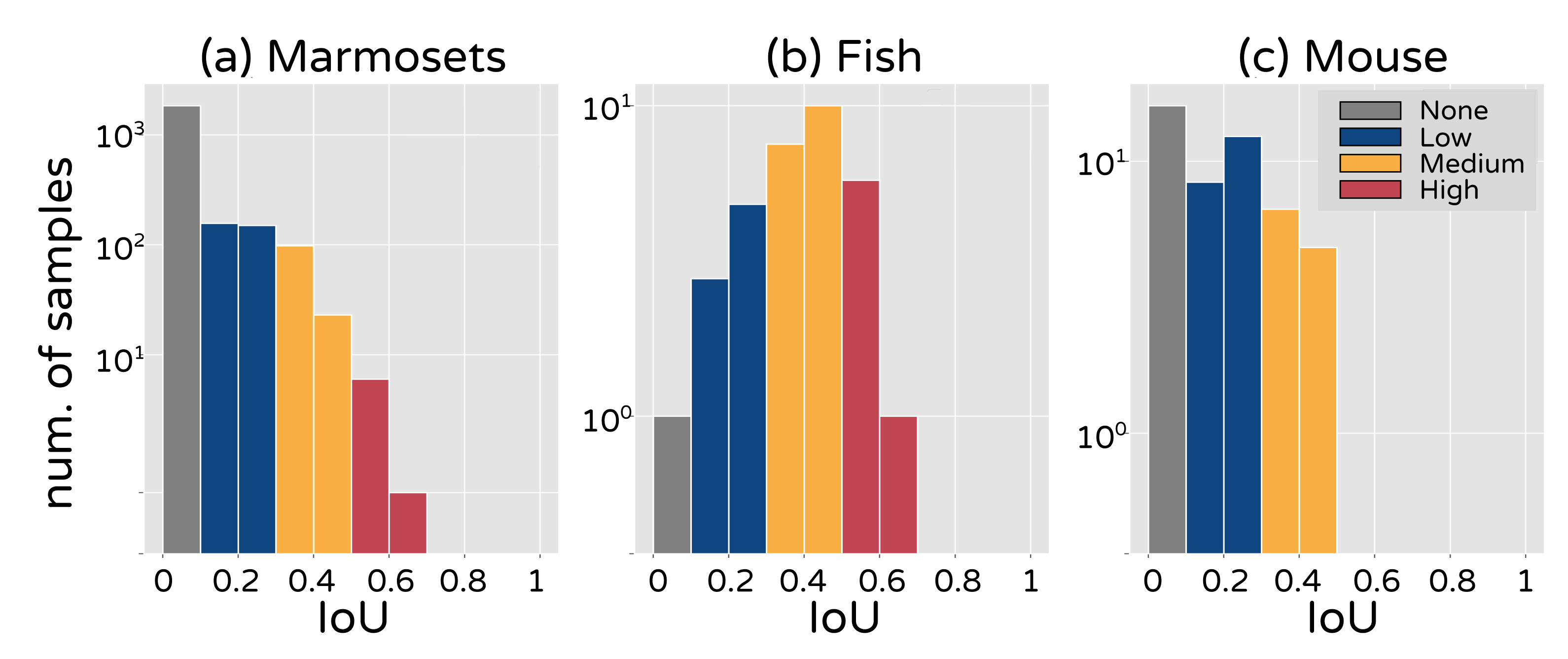}
\end{center} 
\vspace*{-5mm}
\caption{\textbf{Crowdedness levels for the animal datasets}, as divided into low, medium, and hard. The SchoolingFish dataset (Fish) is overall more crowded, while the Tri-Mouse dataset (Mouse) is the least crowded (IoU: Intersection Over Union).}
\label{fig:overlapped_animals}
\vspace*{-10pt}
\end{figure}


\begin{table*}[b!]
\centering
\small
\begin{tabular}{{l|cccc|cccc|cccc}}
\hline 
\multicolumn{1}{c|}{} & \multicolumn{4}{c|}{Marmosets} &  \multicolumn{4}{c|}{SchoolingFish} & \multicolumn{4}{c}{Tri-Mouse} \\
\hline
 Method (detector) & AP & \textcolor{mygreen}{AP$_L$}  & \textcolor{myyellow}{AP$_M$} & \textcolor{myred}{AP$_H$} & AP & \textcolor{mygreen}{AP$_L$}  & \textcolor{myyellow}{AP$_M$} & \textcolor{myred}{AP$_H$}  & AP & \textcolor{mygreen}{AP$_L$}  & \textcolor{myyellow}{AP$_M$} & \textcolor{myred}{AP$_H$} \\
\hline
\multicolumn{2}{l}{Bottom-up methods}  \\
\hline
Resnet~\cite{lauer2022multi}-AE~\cite{newell2017associative} & 45.0 & - & - & -& 40.0 & - & - & - & 70.3 & - & - & -\\
HRNet~\cite{lauer2022multi}-AE~\cite{newell2017associative}& 65.1 & - & - & - &  45.7 & - & - & - &  83.9 & - & - & - 
\\
DLCRNet~\cite{lauer2022multi} & 80.1 & \textcolor{black}{75.3} & \textcolor{black}{71.9} & \textcolor{black}{67.5}  &   74.1 & \textcolor{black}{68.7} & \textcolor{black}{77.6} & \textcolor{black}{72.8} &   95.8 & \textcolor{black}{94.9} & \textcolor{black}{94.7} & - \\
CID-W32~\cite{wang2022contextual} & 92.5  & 90.5  & 91.8  &  82.9 &  81.0  &  72.7 & 84.0  & 79.9  & 86.8 & 84.1  & 85.6  &  -  \\
DEKR~\cite{GengSXZW21} &  61.4  & 64.8   &  64.8  &  57.7 &  77.6 &  82.3  &  80.3 & 72.1  & 97.2 &  97.2  &  97.9  &  -  \\
PETR~\cite{shi2022end} &  93.2  &  89.7  &  89.6  &  70.4  &  79.3  & 72.3   &  71.8  & 80.8  & 82.3 & 84.1  &  78.7 &  -  \\
\hline

\multicolumn{2}{l}{Top-down methods} \\
\hline
HRNet-W48~\cite{wang2020deep} (YOLOv3~\cite{redmon2018yolov3})  & 91.0 & \textcolor{black}{90.1} & \textcolor{black}{ 87.7} & \textcolor{black}{44.1} & 82.9 & \textcolor{black}{80.6}  & \textcolor{black}{85.2}  & \textcolor{black}{79.2} & 91.7  & \textcolor{black}{91.2} & \textcolor{black}{89.5} & - \\
HRNet-W48~\cite{wang2020deep} (Faster R-CNN~\cite{ren2015faster})  & 91.6 & \textcolor{black}{90.2} & \textcolor{black}{85.0} & \textcolor{black}{42.2} & \bf89.1 & \textcolor{black}{82.8} & \textcolor{black}{{\bf92.6}} & \textcolor{black}{86.1}  & 96.0  & \textcolor{black}{97.0} & \textcolor{black}{90.3} & -  \\

\hline
\multicolumn{2}{l}{\textbf{Hybrid approaches}} \\
\hline
BUCTD-preNet-W48 (DLCRNet)  & 90.4 & \textcolor{black}{87.0} & \textcolor{black}{86.1} & \textcolor{black}{{85.7}} & {88.7} &  \textcolor{black}{{\bf85.8}} & \textcolor{black}{90.5} & \textcolor{black}{{88.9}}  &  {98.5}  & \textcolor{black}{ 97.9} & \textcolor{black}{98.3} & - \\
BUCTD-preNet-W48 (CID-W32)  & 93.3  & \bf91.9  & \bf93.4  & \bf89.9  & 88.0 & 79.3  & 90.4  & \bf90.5  & 87.7 & 85.6  & 87.3  &  -  \\
BUCTD-CoAM-W48 (PETR)  & \bf93.7  &  91.3  &  90.6  & 73.8  & 78.8 & 73.8  & 71.9  & 81.0  & 82.7 & 84.1  & 79.5  &  -  \\
BUCTD-CoAM-W48 (DLCRNet)$\sigma$  &   91.6 & \textcolor{black}{ 86.3 } & \textcolor{black}{ 88.9} & \textcolor{black}{89.4} &  81.9  &  \textcolor{black}{{71.0}} & \textcolor{black}{81.0} & \textcolor{black}{78.3}  & \bf99.1 & \textcolor{black}{\bf99.1} & \textcolor{black}{\bf99.2} & - \\
\hline
\end{tabular} 
\vspace{2pt}
\caption{\textbf{BUCTD performance on Animal Pose Datasets.} BUCTD model largely outperforms top-down methods in crowded scenes. Model in brackets denotes where conditions are coming from during inference. Here, $\sigma$ denotes the generative sampling training scheme, while others used empirical sampling is based on DLCRNet predictions.
}
\label{table:table_results_animals}
\end{table*}
These datasets do not contain ground-truth bounding boxes. Thus, we trained the TD models in the same way as CTD models, i.e. bounding boxes of train and test samples were computed from the predictions of DLCRNet. To compare the classic TD pipeline with an object detector, we also ran the TD models in a traditional way: training with ground truth bounding boxes (derived from GT poses), and testing the model with the bounding boxes from the object detectors (Faster R-CNN~\cite{ren2015faster} and YOLOv3~\cite{redmon2018yolov3}, as on CrowdPose). We evaluated BUCTD-W32, -W48 and compared them to TD and BU methods. 

\textbf{Evaluation for different crowdedness levels.} These datasets vary in crowdedness (Figure~\ref{fig:overlapped_animals}). We calculated the Intersection Over Union (IoU) of the overlapped bounding boxes in the images and use the maximum IoU for each image (i.e., maxIoU) to indicate the crowdedness level. We then further split the test set into different crowdedness levels for the test set: low (S$_{L}$), medium (S$_{M}$), and high (S$_{H}$); Figure \ref{fig:overlapped_animals}) to compute the metrics AP$_{L}$, AP$_{M}$, AP$_{H}$, respectively, to interrogate the model performance for different crowdedness levels test set (Table \ref{table:table_results_animals}).

\begin{figure}[ht]
\begin{center}
\setlength{\abovecaptionskip}{0.1cm}
\includegraphics[width=0.48\textwidth]{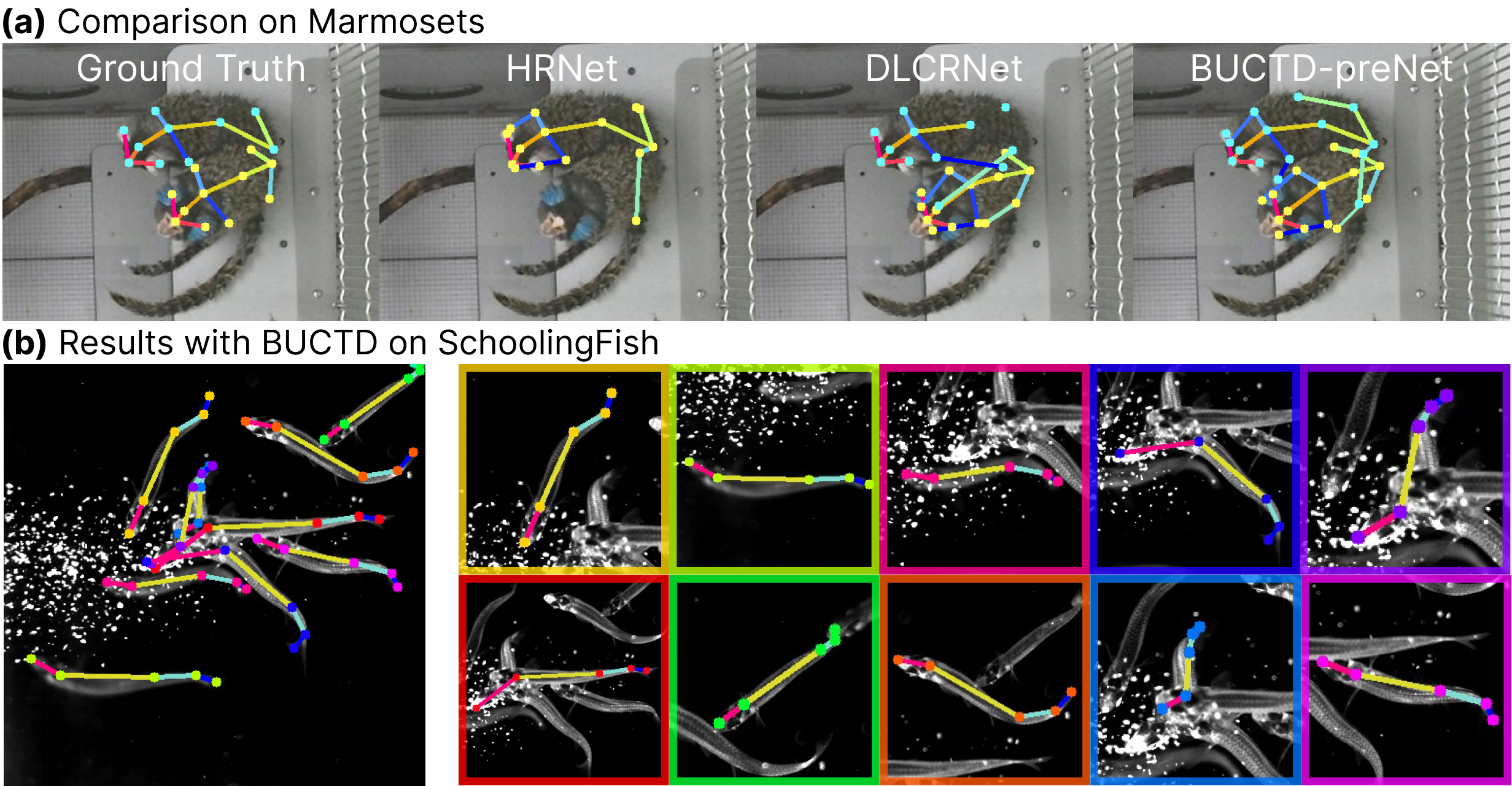}
\end{center} 
\vspace*{-3mm}
\caption{\textbf{Qualitative results on animal datasets.} (a) Each image with two marmosets shows the GT pose, HRNet-W48 predictions, DLCRNet predictions and BUCTD-preNet-W48 predictions (left to right). Due to crowdedness and occlusions, HRNet-W48 often missed the whole individual. DLCRNet may assign the body part to the wrong individual. BUCTD approach shows better performance in highly crowded conditions. (b) Results with BUCTD-preNet-W48 on one zoomed in fish image.}
\label{fig:animal_comparison}
\end{figure}

\textbf{Results:} We trained BUCTD models using the predictions from a bottom-up method based on DeepLabCut's DLCRNet~\cite{lauer2022multi}. BUCTD outperforms the baseline methods on mice and marmosets and is competitive for fish. Importantly, BUCTD outperforms all other methods for crowded frames (see $AP_H$, Table \ref{table:table_results_animals}, and Figure \ref{fig:animal_comparison}). 

Additionally, we trained the CTD models using the generative sampling scheme. During testing, we use the same pose inputs as for BUCTD models. These models also improve the performance on the three animal benchmarks (Table ~\ref{table:table_results_animals}). However, compared to sampling the conditions from BU predictions (empirical sampling), this sampling strategy only performs well for marmosets and mice. We speculate that this is due to the different body plans of mammals and fish.

\textbf{Ablation Results:} One key question is if BUCTD is better because it is based on BU as a detector vs. a standard detector or due to its pose-refinement ability. We find that for crowded scenes, BU detections are key. We fed the pose predictions from top-down methods with standard detectors (YOLOv3, Faster R-CNN), to different CTD models and found that CTD can further improve the performance (Table~\ref{table:error_correction_animal}). This validates that they can refine poses (akin to Table~\ref{table:table_results_crowdpose_errcorr} for CrowdPose). Crucially, when considering the most crowded test data (i.e. $AP_H$ for fish and marmosets and $AP_M$ for mice) then the results were substantially worse than for the full BUCTD pipeline. For example, on the difficult marmoset frames detector with CTD only reached 52.4 mAP vs. 89.4 with BUCTD-CoAM-W48 (DLCRNet). The same is true for other animals and models (Table~\ref{table:error_correction_animal}, and Figure~\ref{fig:fig1}).  

Next, to validate our design choices on the bounding boxes during training, we performed ablation studies for the BUCTD-CoAM-W48 model using the animal benchmarks. The performance of top-down methods is influenced by the quality of the object detection outputs, i.e., the bounding boxes. Different from typical top-down methods, which train the model based on ground truth bounding boxes, and test on detected bounding boxes, we trained the BUCTD model using bounding boxes computed from bottom-up predictions. Validating this design, models trained on bottom-up-computed boxes perform the best on bottom-up-computed boxes during testing (Table \ref{table:table_results_animal_ablations_bbox}).

\begin{table}[t]
\centering
\footnotesize
\begin{tabular}{l|cc|cc|cc}
\hline 
\multicolumn{1}{c|}{} & \multicolumn{2}{c|}{Marmosets} &  \multicolumn{2}{c|}{SchoolingFish} & \multicolumn{2}{c}{Tri-Mouse} \\
\hline
 Method & AP & \textcolor{myred}{AP$_H$} & AP & \textcolor{myred}{AP$_H$}  & AP & \textcolor{myyellow}{AP$_M$} \\
\hline
HRNet-W48 (YOLOv3)  & 91.0 &  44.1 &  82.9 &  79.2 &  91.7 &  89.5\\
BUCTD-CoAM-W32 & 91.3 & 45.5 & \bf87.8 & \bf85.0 & 94.9 & 91.5\\
BUCTD-CoAM-W48 & 91.3 & 48.0 & 86.3 & 82.1 & 92.4 & 90.5\\
BUCTD-preNet-W48 & 91.8 & 50.4 &  85.2 &  79.5 &  91.8 &  89.4\\
BUCTD-CoAM-W48$\sigma$  & \bf93.1 & \bf52.4 & 79.5 & 74.0 & \bf96.9& \bf93.8\\
\hline
HRNet-W48 (FasterRCNN)  & 91.6 &  42.2 &  89.1 &  86.1 &  96.0 &  90.3\\
BUCTD-CoAM-W32 & 91.8 & 42.7 & 90.8 & 88.0 & 96.4 & 90.6\\
BUCTD-CoAM-W48 & 91.6 & 44.8 & \bf90.9 & \bf88.7 & 96.1 &  89.9\\
BUCTD-preNet-W48 &  91.8 & 44.8 & 89.3 & 87.4 & 96.3 & 91.9\\
BUCTD-CoAM-W48$\sigma$  & \bf92.8 & \bf49.8 & 85.0 & 81.0 & \bf97.3 & \bf92.5 \\

\hline
DLCRNet & 80.1 & 67.5 & 74.1 & 72.8 & 95.8 & 94.7 \\
BUCTD-CoAM-W32  &  89.5  & \textcolor{black}{84.3} & 86.9  &  \textcolor{black}{86.5}  & 98.4 & \textcolor{black}{97.6} \\
BUCTD-CoAM-W48  &  89.5  & \textcolor{black}{84.5} & 88.2  & \textcolor{black}{86.6}  & 98.5 & \textcolor{black}{98.3} \\
BUCTD-preNet-W48  & 90.4 & \textcolor{black}{{85.7}} & \bf{88.7} & \textcolor{black}{{\bf88.9}}  &  {98.5} & \textcolor{black}{98.3}\\
BUCTD-CoAM-W48$\sigma$  &  \bf91.6 & \textcolor{black}{\bf89.4} &  81.9  & \textcolor{black}{78.3}  & 
\bf99.1 & \textcolor{black}{\bf99.2}\\

\hline
CID-W32 & 92.5 &  82.9 &  81.0  &  79.9  & 86.8 & 85.6 \\
BUCTD-CoAM-W32 & 93.1  &  84.0 &  86.3 & 85.4  & 90.9  & 88.0  \\
BUCTD-CoAM-W48  & 92.9  & 86.3  &  87.2 & 83.8  & 90.2  & 88.1\\
BUCTD-preNet-W48  & \bf93.3 & 89.9  & \bf88.0  & \bf90.5  & 87.7  & 87.3 \\
BUCTD-CoAM-W48$\sigma$  & 91.8  &  \bf90.9  & 83.2 & 81.1  & \bf94.1 &  \bf92.0 \\
\hline 
\end{tabular} 
\vspace{2pt}
\caption{\textbf{Generalization results on animal benchmarks.} All CTD models boosted predictions from different models provided as conditional inputs. BUCTD models trained on DLCRNet predictions. $\sigma$ denotes generative sampling.}
\label{table:error_correction_animal}
\end{table}

\setlength{\tabcolsep}{1pt}
\begin{table}[H]
\centering
\footnotesize
\begin{tabular}{l|c|c|ccc}
\hline 
\multicolumn{1}{l|}{}&\multicolumn{2}{c|}{Training box}  & \multicolumn{1}{c}{Marmosets} &  \multicolumn{1}{c}{SchoolingFish} & \multicolumn{1}{c}{Tri-Mouse} \\
\hline
Method & GT & BU & AP  & AP  & AP  \\
\hline
HRNet-W48 & \checkmark & - & 85.2  & 73.0  & 97.5 \\
\hline
HRNet-W48 & - & \checkmark &  \bf87.8  & \bf76.0  & \bf98.0 \\
\hline
BUCTD-CoAM-W48 & \checkmark & - & 85.8  &  78.9 &  98.3\\
\hline
BUCTD-CoAM-W48 & - & \checkmark &  \bf90.4  & \bf88.7  &  \bf98.5 \\
\hline
\end{tabular}
\vspace{2pt}
\caption{\textbf{Effect of bounding boxes on training.} Using bottom-up-computed (BU) boxes during training makes models more robust to crops seen during inference. This effect is particularly large for the BUCTD pipeline, validating the choice of training on bounding boxes that are computed with BU predictions (empirical sampling) or GT + error (generative sampling).}
\label{table:table_results_animal_ablations_bbox}
\end{table}

\section{Discussion}
\label{sec:discussion}

Humans and other animals often interact closely making monocular pose estimation a challenging task. The field is increasingly creating benchmarks that encompass these challenges~\cite{li2019crowdpose,zhang2019pose2seg,lauer2022multi}, and new solutions for accurately estimating poses in crowded scenes are actively being developed. 
Here, we presented a new hybrid approach to improve multi-instance pose estimation especially in crowded scenarios, which we call BUCTD.
We compared BUCTD against classic bottom-up, top-down, refinement (PoseFix), and the recent hybrid top-down method (MIPNet), the latter showed excellent performance on crowded frames on challenging human benchmarks. Our BUCTD method achieves state-of-the-art performance on both CrowdPose and OCHuman and new animal pose benchmarks. We believe our proposal of using conditional input provided by a bottom-up pose estimator effectively helps in crowded scenes. 

Of course, there still remains a gap on all the benchmarks considered.
To further enhance performance, future work could additionally leverage hybrid sampling, or model-confidence across keypoints to condition the input predictions. We could also extend BUCTD to tracking of individuals and re-identification across frames. It could potentially also be adapted into video-based architectures, multi-modal learning settings, and with new optimization algorithms~\cite{Chen2023SymbolicDO,bernstein2023automatic}.

In summary, BUCTD is a simple yet effective method that is model-backbone-agnostic. The usage of conditional input provided by a bottom-up pose estimator effectively helps solve heavily occluded and crowded scenes. Thus, we hope it can be a useful addition and broadly applied to newer architectures in computer vision as they arise.

\section*{Acknowledgements}

We are grateful to EPFL and EPFL's School of Life Sciences PTECH fund for providing funding (AM), and the Vallee Foundation (MWM). We thank Steffen Schneider, Shaokai Ye, Haozhe Qi, and other members of the M.W.Mathis Lab and Mathis Group for feedback.

\section*{Supplementary Materials}

\setcounter{section}{0}
\renewcommand{\thesection}{\Alph{section}}
\setcounter{table}{0}
\renewcommand{\thetable}{S\arabic{table}}
\setcounter{figure}{0}
\renewcommand{\thefigure}{S\arabic{figure}}


In the following, we provide more qualitative examples, additional implementation details, compare the general detection performance of the bottom-up stage, and evaluate the type of conditional training data that gives best CTD results. We conclude by showing that the computational costs of BUCTD compare favorably to the standard detector+TD pipeline.

Overall, our work demonstrated that providing individual detections and conditional pose input derived from a bottom-up pose detector to a Conditional Pose Estimator (CTD) can effectively boost performance in crowded scenes.

\begin{figure*}[t]
\begin{center}
\includegraphics[width=\textwidth]{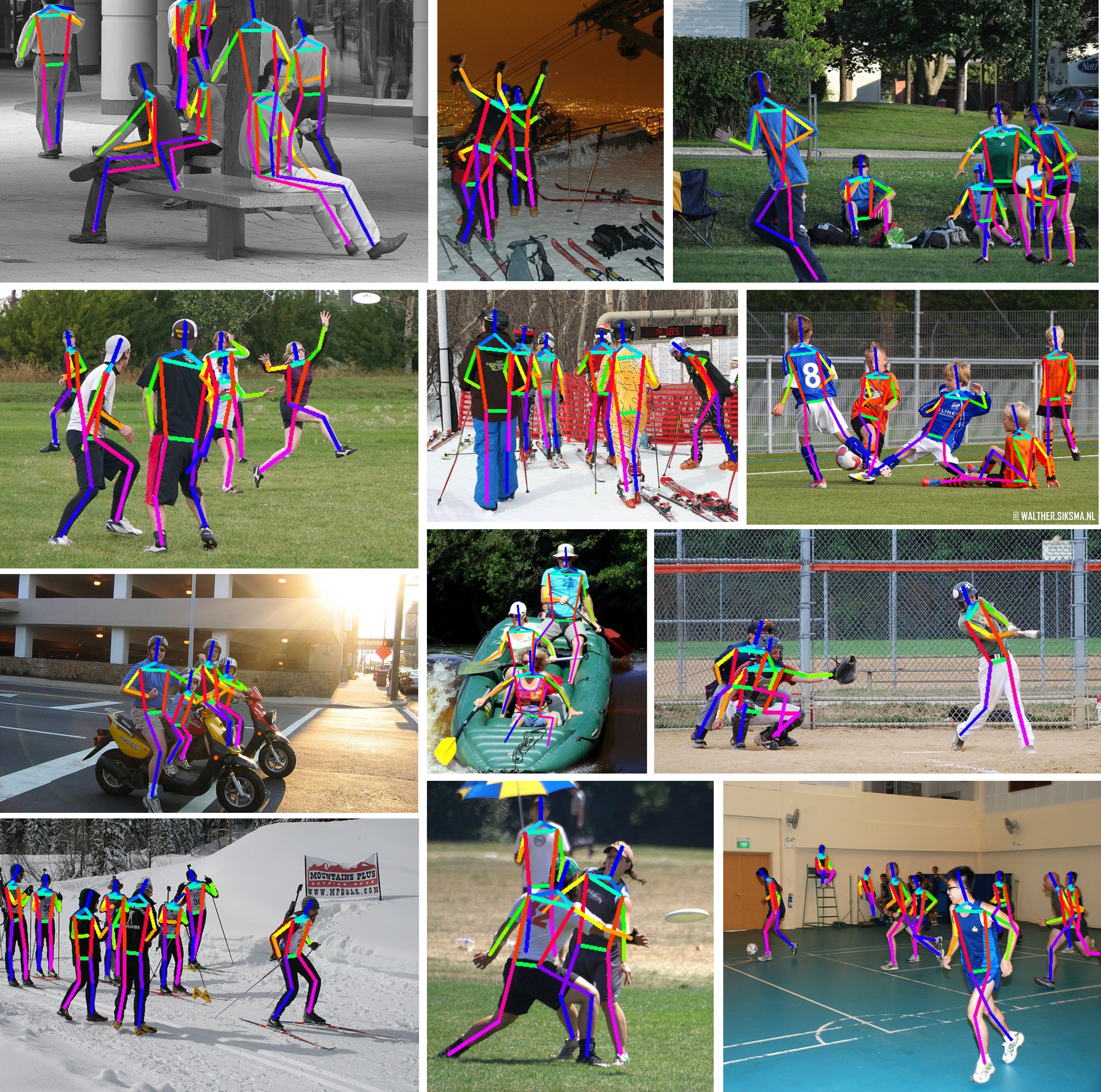}
\end{center} 
\vspace*{-3mm}
\caption{Additional predictions using BUCTD-CoAM-W48 with conditional inputs from PETR on the CrowdPose $test$ set.}
\label{fig:vis_results_2}
\end{figure*}

\section{Success and failure cases}
\label{sec:successandfailure}

\begin{figure*}[t]
\begin{center}
\includegraphics[width=\textwidth]{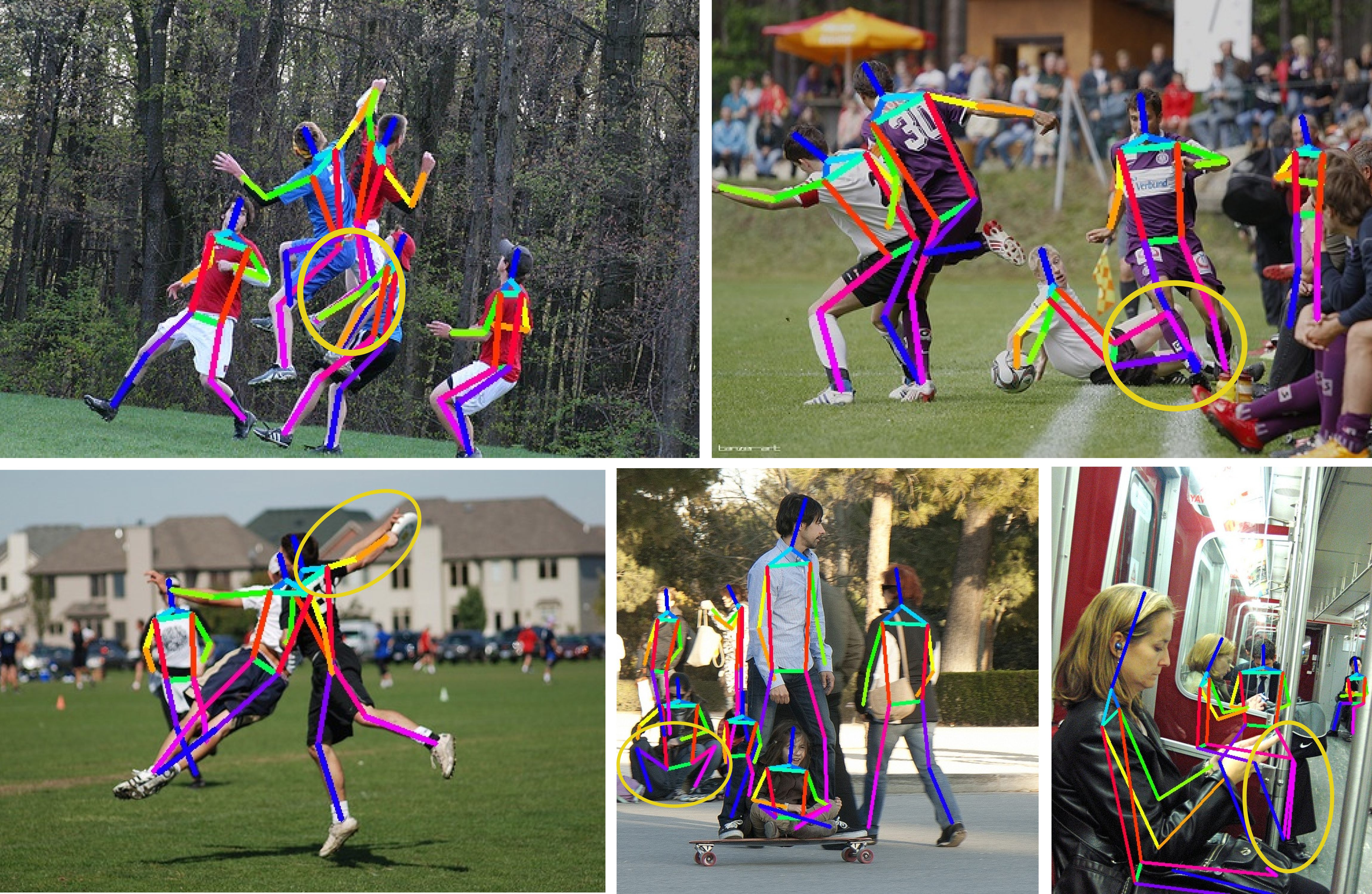}
\end{center} 
\vspace*{-3mm}
\caption{Example failure cases from the CrowdPose $test$ set. Especially in complex sport scenes, our BUCTD-CoAM-W48 model sometimes shows inaccuracies in estimating the correct position of the extremities. Errors are highlighted by yellow ellipses.}
\label{fig:failures}
\end{figure*}

To illustrate the power of our method, we show additional qualitative results of success (Figure~\ref{fig:vis_results_2}) and failure cases (Figure~\ref{fig:failures}).

\section{Implementation details}
\label{sec:implementation_details}
We report training and implementation details for the bottom-up and the conditional top-down models applied to both human benchmarks and animal benchmarks.

\subsection{Training settings for bottom-up pose detectors}

In order to create pose proposals from bottom-up models, we trained DLCRNet-ms4~\cite{lauer2022multi} on the animal benchmarks, and HigherHRNet-W32 using mmpose~\cite{mmpose2020} on the human benchmarks. We show the default training settings in Table \ref{table:training_params_bu_models}.

\begin{table}[ht]
\centering
\footnotesize
\begin{tabular}{l|cc}
\hline
Hyperparameters & Animal Benchmarks & Human Benchmarks \\
 & DLCRNet & HrHRNet-W32 \\
\hline
Optimizer & Adam~\cite{kingma2014adam} & Adam~\cite{kingma2014adam} \\
Base learning rate & 0.0001 & 0.0015 \\
Learning rate sched. &  step [7500, 12000]  & step [200, 260] \\
Learning rate drop ($\gamma$) & [0.5, 0.2] & 0.1 \\
Training epochs & - & 300 \\
Training iterations & 60,000  & - \\
Warmup iterations & - & 500 \\
Warmup ratio & - & 0.001 \\
Batch size & 8 & 40 \\
\hline
Input resolution & 400 $\times$ 400  & 512 $\times$ 512 \\
Rotation & 24\degree & 30\degree \\
Scale & [0.5, 1.25]& [0.75, 1.5] \\
RandomFlip & - & 0.5 \\
\hline
\end{tabular}
\vspace{5pt}
\caption{{\bf Default training settings for bottom-up models.} We applied these hyperparameters and training data settings to the bottom-up models (DLCRNet for animal datasets, and HigherHRNet for human benchmarks).}
\label{table:training_params_bu_models}
\end{table}

Note that the HigherHRNet-W32 is trained with the same settings for COCO~\cite{lin2014microsoft} (used also for inference on OCHuman~\cite{zhang2019pose2seg}) and CrowdPose~\cite{li2019crowdpose}. We furthermore used the same settings of the DLCRNet as in Table~\ref{table:training_params_bu_models} on all three animal datasets.

\subsection{Training settings for conditional top-down (CTD) pose estimators}

Here we provide the default parameter settings for CTD-CoAM-W32/48 and CTD-TP-H-A6 trained on human benchmarks (Table~\ref{table:training_params_ctd_models}). 

\setlength{\tabcolsep}{3.pt}
\begin{table}[ht]
\centering
\footnotesize
\begin{tabular}{l|c|c}
\hline
Hyperparameters &  CTD-CoAM-W32/48 & CTD-TP-H-A6 \\
\hline
Optimizer & Adam~\cite{kingma2014adam} & Adam~\cite{kingma2014adam} \\
Base learning rate & 0.001 & 0.0001 \\
Weight decay & 0.0001 & 0.1 \\
Learning rate sched. & step [170,200] & step [100,150,200,220] \\
Learning rate drop ($\gamma$) & 0.1 & 0.25 \\
Training epochs & 210 & 240 \\
Batch size & 96/48 & 64 \\
\hline
Input resolution & 256$\times$192 / 384$\times$288  & 256$\times$192 \\
Rotation & 45\degree & 45\degree \\
Scale & [0.65, 1.35]  & [0.65, 1.35] \\
RandomFlip & 0.5 & 0.5 \\
\hline
\end{tabular}
\vspace{5pt}
\caption{{\bf Default training settings for conditional top-down models.} We apply these hyperparameters and training data settings to a CTD-CoAM-W32.}
\label{table:training_params_ctd_models}
\end{table}

During training on the animal benchmarks, we used the same settings for CTD-CoAM-W32/W48 (Table~\ref{table:training_params_ctd_models}). However, to keep the aspect ratio for animals, we padded the cropped individuals to the input resolution of 256$\times$256 (and the batch size is 16). For training CTD-CoAM-W32 on COCO~\cite{lin2014microsoft}, we trained for 110 epochs, with an initial learning rate of 0.02 and a learning rate drop at epochs 70 and 110, respectively.


\subsection{Details for CTD with a Conditional Attention Module}
\label{subsec:CoAMdetails}

We introduced a new Conditional Attention Module (CoAM) that is inspired by Fu et al.~\cite{fu2019dual} and contains spatial and channel attention sub-modules, which are defined as follows. 

\textbf{Position Attention Module.}

Given a local feature $f$, we first feed it into a convolution layer to obtain $F \in \mathbb{R}^{C \times H \times W}$ and embed it linearly to generate two new feature maps $K$ and $V$ (keys and values) with ${K,V} \in \mathbb{R}^{C \times H \times W}$. The condition heatmap $c \in \mathbb{R}^{3 \times H \times W}$ at the corresponding stage is also processed by a convolution layer to create $C \in \mathbb{R}^{3 \times H \times W}$ and embedded linearly into $Q$ (queries) with $Q \in \mathbb{R}^{C \times H \times W}$. We reshape queries, keys and values to $\mathbb{R}^{C \times N}$, where $N = H \times W$ is the number of pixels. A softmax layer is applied after a matrix multiplication between the transpose of $Q$ and $K$, to generate the spatial attention map $S \in \mathbb{R}^{N \times N}$:
\begin{equation} \label{eq:pos_att_map}
\setlength{\abovedisplayskip}{3pt}
\setlength{\belowdisplayskip}{3pt}
s_{ji} = \frac{\exp(Q_i \cdot K_j)}{\sum_{i=1}^{N}\exp(Q_i \cdot K_j)}
\end{equation}
where $s_{ji}$ measures the impact of condition position $i$ on the feature position $j$. Then, we perform a matrix multiplication between $V$ and the transpose of $S$ and reshape the result to $\mathbb{R}^{C \times H \times W}$ to obtain the final output $P$ of the position attention submodule:
\begin{equation} \label{eq:pos_att}
\setlength{\abovedisplayskip}{1pt}
\setlength{\belowdisplayskip}{1pt}
P_j = \sum_{i=1}^{N} (s_{ji}V_i)
\end{equation}
The resulting feature from the position attention submodule has a global contextual view and aggregates the conditional context according to the spatial attention map. \\

\textbf{Channel Attention Module.}

Each channel map of high-level features can be regarded as a keypoint-specific response while the condition itself is a keypoint-specific map. Hence, it is beneficial to learn the associations between these different semantic representations.

Different from the position attention submodule, the channel attention submodule directly calculates the channel attention map $X \in \mathbb{R}^{C \times C}$ from the original features $F$ (treated as key and value) and the condition $C$ (processed by convolution layer to be in $\mathbb{R}^{C \times H \times W}$ and treated as query). Specifically, we reshape both $F$ and $C$ to $\mathbb{R}^{C \times N}$, and then perform a matrix multiplication between $F$ and the transpose of $C$, followed by a softmax layer to retain the channel attention map $X \in \mathbb{R}^{C \times C}$:
\begin{equation} \label{eq:pos_att_map2}
\setlength{\abovedisplayskip}{3pt}
\setlength{\belowdisplayskip}{3pt}
x_{ji} = \frac{\exp(C_i \cdot F_j)}{\sum_{i=1}^{C}\exp(C_i \cdot F_j)} 
\end{equation}
where $x_{ji}$ measure the impact of the condition channel $i$ on the feature channel $j$. Afterwards, we perform a matrix multiplication between the transpose of $X$ and $F$ and reshape the result to $\mathbb{R}^{C \times H \times W}$ to obtain the final output $E$ of the channel attention submodule:
\begin{equation} \label{eq:pos_att3}
\setlength{\abovedisplayskip}{1pt}
\setlength{\belowdisplayskip}{1pt}
E_j = \sum_{i=1}^{C} (x_{ji}F_i)
\end{equation}
The final feature of the channel attention submodule models the long-range semantic dependencies between conditional keypoints and feature maps.

To obtain the final output $M$ of CoAM, we perform an element-wise sum operation between the original feature map $F$ and the outputs of the respective submodules $P$ and $E$:
\begin{equation} \label{eq:pos_att4}
\setlength{\abovedisplayskip}{.5pt}
\setlength{\belowdisplayskip}{.5pt}
M_j = F_j + (P_j + E_j)
\end{equation}

\subsection{Design of the conditional input to CTD}
\label{subsec:ValidationHM}

The condition fed to the CTD stage of our BUCTD framework is created as follows:

With the predictions coming from the first stage, we generate a conditional heatmap in ($c \in \mathbf{R}^{H\times W\times 3}$) by using a Gaussian distribution with a standard deviation $\sigma$. 

We tried several designs for this conditional input: (3D) color heatmap (CM), (1D) gray-scale heatmap (GM), and K-channel single Gaussian heatmaps (KM). We achieved +1.3 mAP with CM, +0.9 mAP with GM, vs. KM BUCTD-CoAM-W32 on CrowdPose. 

Therefore, we applied color heatmap as conditions for all models.

\subsection{Details of generative sampling scheme during conditional training}
\label{subsec:generativesampling}

Similar to PoseFix~\cite{moon2019posefix}, during training, we synthesized the pose by using the error statistics described in ~\cite{ruggero2017benchmarking} as prior information to generate noisy pose as conditional inputs. We generated the conditional pose with the four error types of jitter, inversion, swap and miss. For human benchmark, i.e. CrowdPose, we applied the same error probabilities as in PoseFix (which are estimated from COCO and are likely slightly different~\cite{ruggero2017benchmarking}; despite this we achieve excellent results). For animal benchmarks, we utilized the same error types and tuned the error distribution by running a few different cases; we ended up using jitter error: 0.15 or 0.2 (depending on keypoint validity), miss error: 0.05 or 0.2 (depending on keypoint validity), inversion error: 0.03, swap error: 0.04 or 0.1(depending on keypoint validity). Additionally, we allow swapping keypoints between individuals that do not have any overlap, to simulate wrong assemblies in the bottom-up stage. 

Our results demonstrate that the CoAM module leads to improved performance on some animal benchmarks when applying generative sampling (Table \ref{table:generative_sampling_animals}). However, the preNet module underperforms on the SchoolingFish dataset compared to the baseline results. We further ablate the error types and find that the performance with fewer error types on preNet-W48 on the SchoolingFish dataset is slightly higher than the performance on the models with all error types. Specifically, when we use two types of errors (jitter and swap), we achieve 71.7 AP, while using jitter error only results in 77.0 AP. 

From the different results, we observed that the generative sampling strategy is not as stable as empirical sampling on small-scale datasets, likely due to different error statistics between human and animal pose estimation methods (or different body plans). However, combined generative and empirical sampling could be a great strategy to explore in the future.

\begin{table}[ht]
\centering
\footnotesize
\begin{tabular}{l|ccc}
\toprule
methods & Marmosets & Sch.Fish & Tri-Mouse \\
\midrule
BUCTD-preNet-W48 (DLCRNet) & 91.6 & 62.1 & 98.4\\
BUCTD-CoAM-W48 (DLCRNet) & 91.6 & 81.9 & 99.1 \\
BUCTD-preNet-W48 (DLCRNet)$\sigma$ & 90.4 & 88.7 & 98.5  \\
\bottomrule
\end{tabular}
\vspace{5pt}
\caption{Results on animal benchmarks with generative sampling and empirical sampling. $\sigma$ denotes empirical sampling.}
\label{table:generative_sampling_animals}
\end{table}

\section{Further Comparisons to MIPNet}

In this section, we compare our method with previous SOTA (MIPNet).  based on precision and recall. We find higher performance on both metrics with BUCTD, also strong error-correcting capabilities to improve the performance of BU models. 

Furthermore, to ablate the influence of the number of detections (which vary widely across different BU models), we only provide the same (amount of) detected bounding boxes as in MIPNet to our CTD models. We notice that our method still outperforms MIPNet, independent of the bottom-up model applied, and with especially large gains on hard frames (i.e. frames with higher crowdedness level).

\subsection{Evaluation on ground-truth bounding boxes}
\label{sec:eval_gt_bbox}

First, to take the detectors completely out of the equation, we simply evaluated different models on ground truth bounding boxes (i.e., the same pixel input). 

We compare the performance of our BUCTD-CoAM-W32 model on CrowdPose to HRNet and MIPNet when evaluated using ground-truth bounding boxes (Table~\ref{table:table_results_crowdpose_gt}). Note that these models were trained on $train$ and validated on $val$ (as done in~\cite{khirodkar2021multi}). During training we matched the conditions to the GT keypoints and then fed it to the CTD model together with the cropped input image. The same approach is used during testing. Our method outperforms the HRNet baseline and improves upon the MIPNet baseline, that was designed to better handle crowded scenarios. While MIPNet only achieves small improvements over HRNet, our method substantially boosts the AP values, especially on the hard, highly crowded cases (+ 9.0 AP over HRNet and + 6.9 AP over MIPNet). This directly corroborates our choice to provide conditional pose input to boost performance vs. an index.

\begin{table}[ht!]
\centering
\small
\begin{tabular}{l|c|ccc}
\toprule
Method & AP & AP$_{easy}$ & AP$_{med}$ & AP$_{hard}$ \\
\toprule
HRNet-W32~\cite{wang2020deep} & 70.0 & 78.8 & 70.3 & 61.7 \\
MIPNet-W32~\cite{khirodkar2021multi} & 71.2 & 78.8 & 71.5 & 63.8 \\
{\bf BUCTD-CoAM-W32 (Ours)} & {\bf 75.2} & {\bf 81.4} & {\bf 75.3} & {\bf 70.7} \\
\bottomrule
\end{tabular}
\vspace{5pt}
\caption{Our BUCTD model outperforms HRNet and MIPNet on CrowdPose $val$ (using ground-truth bounding boxes). All models are trained on input resolutions of 256x192.}
\label{table:table_results_crowdpose_gt}
\end{table}

\subsection{Performance details - precision and recall}
\label{subsec:precisionrecall}

\begin{figure}[ht]
\begin{center}
\includegraphics[width=0.48\textwidth]{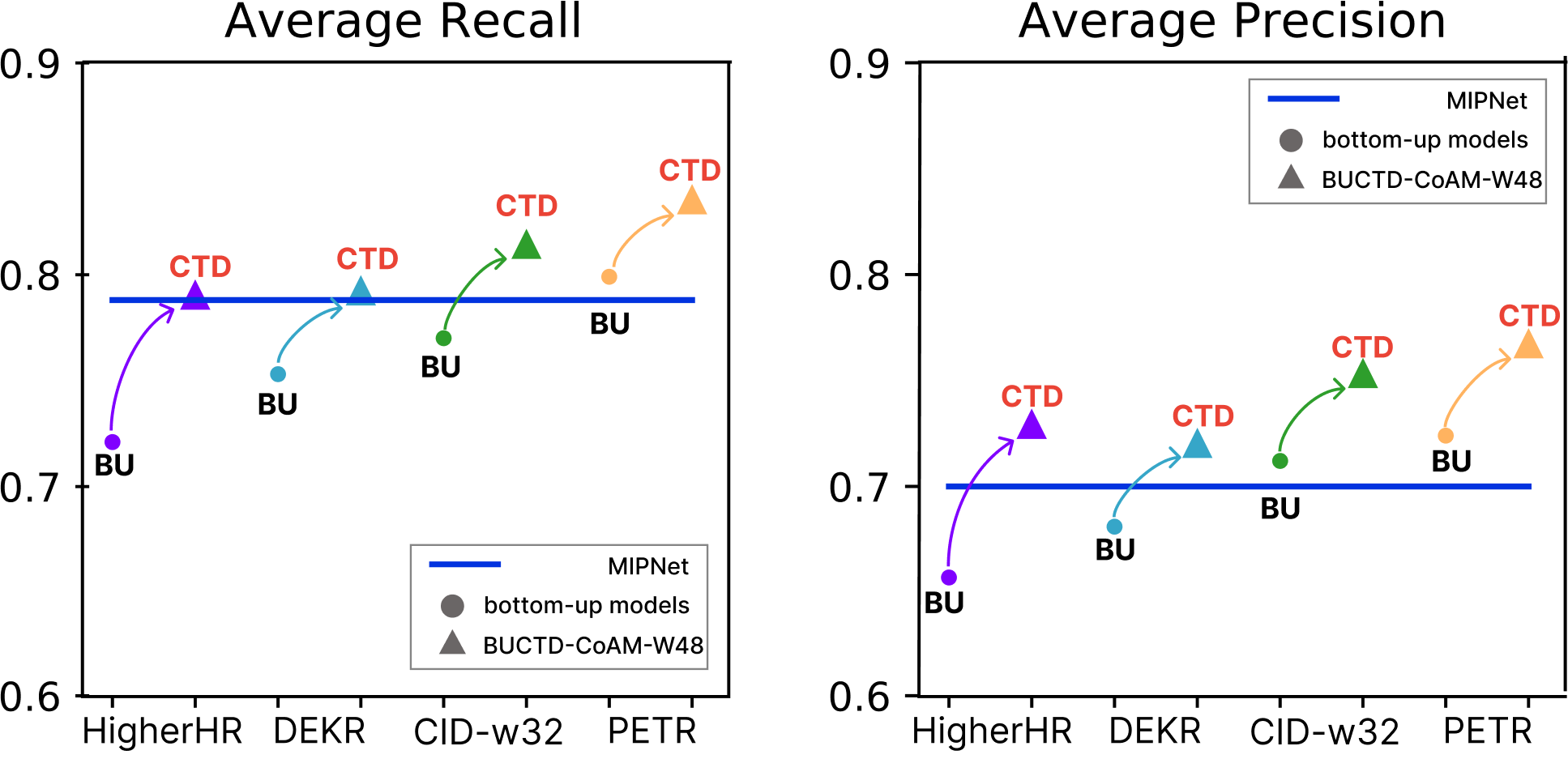}
\end{center} 
\vspace*{-3mm}
\caption{Comparison of recall and precision curves for different BU and CTD models vs. MIPNet. CTD boosts both precision and recall (of BU models), and can thus "recover" more poorly predicted persons than MIPNet.}
\label{fig:recall_precision}
\end{figure}

To gain better insights into the performance gains of BUCTD, we computed precision and recall on the CrowdPose $test$ set (Figure~\ref{fig:recall_precision}). We compared our model (trained with empirical sampling) to the previous SOTA on CrowdPose: MIPNet. Importantly, we have higher recall and precision than MIPNet for all BU models. Thus, due to its design, BUCTD improves the precision \textit{and recall} for all BU models we tested.

\subsection{Robustness to number of detections}
\label{subsec:ablation-detections}

Next, we wanted to fairly compare our BUCTD in terms of the
number of detections that the first stage provides, in order to exclude that simply a higher number of detections, made by the bottom-up pose detector in comparison to commonly used object detectors, would lead to our superior performance.

\begin{figure}[ht]
\begin{center}
\includegraphics[width=0.5\textwidth]{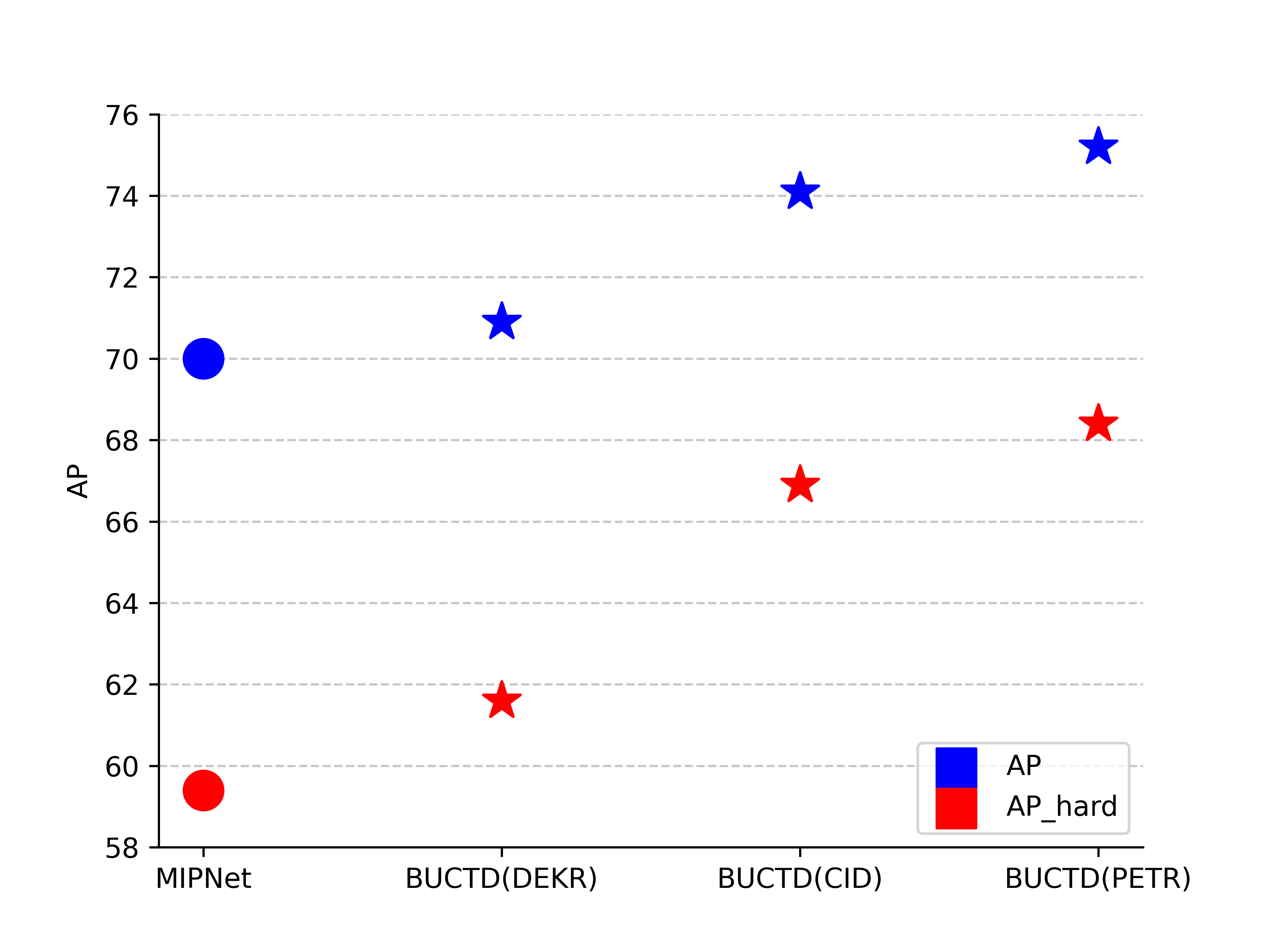}
\end{center} 
\vspace*{-3mm}
\caption{\textbf{Performance of BUCTD when provided with same amount of detections (65,044) as the object detector of MIPNet.} 
When constraining the number of detections we pass from the bottom-up pose detector (DEKR, CID, PETR) to the CTD model to the same amount of detections MIPNet receives from the object detector on the CrowdPose test set, the BUCTD framework still significantly outperforms MIPNet on AP and $AP_H$ (i.e., AP hard).}
\label{fig:sameNrDets}
\end{figure}

We hence provided the \textit{same number} of detections from the bottom-up models, as provided by the object detector. Despite this artificial constraint the performance of BUCTD was still significantly higher than the one of MIPNet~\cite{khirodkar2021multi} (Figure~\ref{fig:sameNrDets}).

\section{Computational costs of training BUCTD}
\label{sec:compcosts}

There are three components for creating a BUCTD model (and comparing it to the standard pipeline either BU or detector + TD). Naturally, simply using a BU model is more efficient, but we also achieve more accuracy. 

In contrast, BU models are more efficient than detectors (as we show below). Creating and storing the empirical predictions as well as matching them to ground-truth also comes with some cost (while one does not have these costs with generative sampling; however, one might need to estimate the error distribution). Training CTD or TD models is comparable. Furthermore, the inference of TD and CTD is also comparable (with an advantage for TD). However, those costs come at the benefit of stronger performance -- depending on the application performance or speed might be differently relevant.

We compared the parameters and GFLOPs on object detectors, bottom-up models, top-down models, and our methods (Table~\ref{table:computational_costs}). Bottom-up models generally have fewer parameters and GFLOPs than widely-used object detectors. 

We further compare the overall training time of object detectors vs. bottom-up pose detectors. We trained two commonly used object detectors (i.e., Faster R-CNN~\cite{ren2015faster} and YOLOv3~\cite{redmon2018yolov3}) with the default parameter settings from mmdetection~\cite{mmdetection} and a DLCRNet~\cite{lauer2022multi} as the bottom-up pose detector on the animal benchmarks. For training on the marmosets datasets, training the FasterRCNN detector with 90 epochs took 26.5 hours (saturated around 60 epochs), and training the YOLOv3 with 273 epochs took 33 hours (saturated around 258 epochs). However, training the DLCRNet took only 2.25 hours for 90 epochs. Experiments were performed on a single Titan RTX. 


\begin{table}[ht!]
\centering
\begin{tabular}{l|cc}
\toprule
Method  & \#params & GFLOPs  \\
\hline
FasterRCNN~\cite{ren2015faster} & 60.0M & 246.0 \\
YOLOv3~\cite{redmon2018yolov3} & 62.0M & 65.9\\
\hline
HigherHRnet-W32~\cite{cheng2020higherhrnet} & 28.6M & 47.9 \\
DEKR~\cite{GengSXZW21} & 28.6M & 44.5 \\
CID~\cite{wang2022contextual} & 29.4M & 43.2 \\
PETR~\cite{shi2022end} & 220.5M & - \\
\hline
HRNet-W48~\cite{khirodkar2021multi}  & 63.6M & 19.5  \\
MIPNet-W48~\cite{wang2020deep} & 63.7M & 64.5 \\
PoseFix~\cite{moon2019posefix} & 68.7M & 36.6 \\
\midrule
BUCTD-preNet-W32 & 28.5M & 7.6 \\
BUCTD-TP-H-A6 & 17.0M & 8.4  \\
BUCTD-CoAM-W32 & 39.1M & 8.6 \\
BUCTD-CoAM-W48  & 115.6M & 43.5 \\
\bottomrule
\end{tabular}
\vspace{5pt}
\caption{Number of parameters and GFLOPs on object detectors, bottom-up models, top-down models and our methods.}
\label{table:computational_costs}
\end{table}

\newpage

\bibliographystyle{ieee_fullname}

\begin{thebibliography}{10}\itemsep=-1pt

\bibitem{andriluka14cvpr}
Mykhaylo Andriluka, Leonid Pishchulin, Peter Gehler, and Bernt Schiele.
\newblock 2d human pose estimation: New benchmark and state of the art
  analysis.
\newblock In {\em IEEE Conference on Computer Vision and Pattern Recognition
  (CVPR)}, June 2014.

\bibitem{artacho2021bapose}
Bruno Artacho and Andreas Savakis.
\newblock Bapose: Bottom-up pose estimation with disentangled waterfall
  representations.
\newblock {\em arXiv preprint arXiv:2112.10716}, 2021.

\bibitem{bernstein2023automatic}
Jeremy Bernstein, Chris Mingard, Kevin Huang, Navid Azizan, and Yisong Yue.
\newblock Automatic gradient descent: Deep learning without hyperparameters,
  2023.

\bibitem{braso2021center}
Guillem Bras{\'o}, Nikita Kister, and Laura Leal-Taix{\'e}.
\newblock The center of attention: Center-keypoint grouping via attention for
  multi-person pose estimation.
\newblock In {\em Proceedings of the IEEE/CVF International Conference on
  Computer Vision}, pages 11853--11863, 2021.

\bibitem{cai2019exploiting}
Yujun Cai, Liuhao Ge, Jun Liu, Jianfei Cai, Tat-Jen Cham, Junsong Yuan, and
  Nadia~Magnenat Thalmann.
\newblock Exploiting spatial-temporal relationships for 3d pose estimation via
  graph convolutional networks.
\newblock In {\em ICCV}, pages 2272--2281, 2019.

\bibitem{cao2019openpose}
Zhe Cao, Gines Hidalgo, Tomas Simon, Shih-En Wei, and Yaser Sheikh.
\newblock Openpose: Realtime multi-person 2d pose estimation using part
  affinity fields, 2019.

\bibitem{mmdetection}
Kai Chen, Jiaqi Wang, Jiangmiao Pang, Yuhang Cao, Yu Xiong, Xiaoxiao Li,
  Shuyang Sun, Wansen Feng, Ziwei Liu, Jiarui Xu, Zheng Zhang, Dazhi Cheng,
  Chenchen Zhu, Tianheng Cheng, Qijie Zhao, Buyu Li, Xin Lu, Rui Zhu, Yue Wu,
  Jifeng Dai, Jingdong Wang, Jianping Shi, Wanli Ouyang, Chen~Change Loy, and
  Dahua Lin.
\newblock {MMDetection}: Open mmlab detection toolbox and benchmark.
\newblock {\em arXiv preprint arXiv:1906.07155}, 2019.

\bibitem{Chen2023SymbolicDO}
Xiangning Chen, Chen Liang, Da Huang, Esteban Real, Kaiyuan Wang, Yao Liu, Hieu
  Pham, Xuanyi Dong, Thang Luong, Cho-Jui Hsieh, Yifeng Lu, and Quoc~V. Le.
\newblock Symbolic discovery of optimization algorithms.
\newblock {\em ArXiv}, abs/2302.06675, 2023.

\bibitem{chen2020alphatracker}
Zexin Chen, Ruihan Zhang, Yu~Eva Zhang, Haowen Zhou, Hao-Shu Fang, Rachel~R
  Rock, Aneesh Bal, Nancy Padilla-Coreano, Laurel Keyes, Kay~M Tye, et~al.
\newblock Alphatracker: a multi-animal tracking and behavioral analysis tool.
\newblock {\em biorxiv}, 2020.

\bibitem{cheng2020higherhrnet}
Bowen Cheng, Bin Xiao, Jingdong Wang, Honghui Shi, Thomas~S Huang, and Lei
  Zhang.
\newblock Higherhrnet: Scale-aware representation learning for bottom-up human
  pose estimation.
\newblock {\em Proceedings of the IEEE/CVF conference on computer vision and
  pattern recognition}, pages 5386--5395, 2020.

\bibitem{cheng2022dual}
Yu Cheng, Bo Wang, and Robby~T Tan.
\newblock Dual networks based 3d multi-person pose estimation from monocular
  video.
\newblock {\em IEEE Transactions on Pattern Analysis and Machine Intelligence},
  45(2):1636--1651, 2022.

\bibitem{mmpose2020}
MMPose Contributors.
\newblock Openmmlab pose estimation toolbox and benchmark.
\newblock \url{https://github.com/open-mmlab/mmpose}, 2020.

\bibitem{ding20222r}
Yiwei Ding, Wenjin Deng, Yinglin Zheng, Pengfei Liu, Meihong Wang, Xuan Cheng,
  Jianmin Bao, Dong Chen, and Ming Zeng.
\newblock I{\({^2}\)}r-net: Intra- and inter-human relation network for
  multi-person pose estimation.
\newblock In Luc~De Raedt, editor, {\em Proceedings of the Thirty-First
  International Joint Conference on Artificial Intelligence, {IJCAI} 2022,
  Vienna, Austria, 23-29 July 2022}, pages 855--862. ijcai.org, 2022.

\bibitem{yolov5}
Glenn~Jocher et. al.
\newblock {ultralytics/yolov5: v6.0 - YOLOv5n 'Nano' models, Roboflow
  integration, TensorFlow export, OpenCV DNN support}.
\newblock {\em Zenodo}, Oct. 2021.

\bibitem{fang2018rmpe}
Hao-Shu Fang, Shuqin Xie, Yu-Wing Tai, and Cewu Lu.
\newblock Rmpe: Regional multi-person pose estimation, 2018.

\bibitem{fieraru2018learning}
Mihai Fieraru, Anna Khoreva, Leonid Pishchulin, and Bernt Schiele.
\newblock Learning to refine human pose estimation.
\newblock In {\em Proceedings of the IEEE conference on computer vision and
  pattern recognition workshops}, pages 205--214, 2018.

\bibitem{fu2019dual}
Jun Fu, Jing Liu, Haijie Tian, Yong Li, Yongjun Bao, Zhiwei Fang, and Hanqing
  Lu.
\newblock Dual attention network for scene segmentation.
\newblock In {\em Proceedings of the IEEE/CVF conference on computer vision and
  pattern recognition}, pages 3146--3154, 2019.

\bibitem{GengSXZW21}
Zigang Geng, Ke Sun, Bin Xiao, Zhaoxiang Zhang, and Jingdong Wang.
\newblock Bottom-up human pose estimation via disentangled keypoint regression.
\newblock In {\em Proceedings of the IEEE/CVF Conference on Computer Vision and
  Pattern Recognition}, 2021.

\bibitem{he2017mask}
Kaiming He, Georgia Gkioxari, Piotr Doll{\'a}r, and Ross Girshick.
\newblock Mask r-cnn.
\newblock In {\em Proceedings of the IEEE international conference on computer
  vision}, pages 2961--2969, 2017.

\bibitem{hu2016bottom}
Peiyun Hu and Deva Ramanan.
\newblock Bottom-up and top-down reasoning with hierarchical rectified
  gaussians.
\newblock In {\em CVPR}, pages 5600--5609, 2016.

\bibitem{InsafutdinovAPT16}
Eldar Insafutdinov, Mykhaylo Andriluka, Leonid Pishchulin, Siyu Tang, Evgeny
  Levinkov, Bjoern Andres, and Bernt Schiele.
\newblock Articulated multi-person tracking in the wild.
\newblock {\em CoRR}, abs/1612.01465, 2016.

\bibitem{jin2020differentiable}
Sheng Jin, Wentao Liu, Enze Xie, Wenhai Wang, Chen Qian, Wanli Ouyang, and Ping
  Luo.
\newblock Differentiable hierarchical graph grouping for multi-person pose
  estimation.
\newblock In {\em European Conference on Computer Vision}, pages 718--734.
  Springer, 2020.

\bibitem{khirodkar2021multi}
Rawal Khirodkar, Visesh Chari, Amit Agrawal, and Ambrish Tyagi.
\newblock Multi-instance pose networks: Rethinking top-down pose estimation.
\newblock In {\em ICCV}, 2021.

\bibitem{kingma2014adam}
Diederik~P Kingma and Jimmy Ba.
\newblock Adam: A method for stochastic optimization.
\newblock {\em arXiv preprint arXiv:1412.6980}, 2014.

\bibitem{lauer2022multi}
Jessy Lauer, Mu Zhou, Shaokai Ye, William Menegas, Steffen Schneider, Tanmay
  Nath, Mohammed~Mostafizur Rahman, Valentina Di~Santo, Daniel Soberanes,
  Guoping Feng, et~al.
\newblock Multi-animal pose estimation, identification and tracking with
  deeplabcut.
\newblock {\em Nature Methods}, 19(4):496--504, 2022.

\bibitem{li2019crowdpose}
Jiefeng Li, Can Wang, Hao Zhu, Yihuan Mao, Hao-Shu Fang, and Cewu Lu.
\newblock Crowdpose: Efficient crowded scenes pose estimation and a new
  benchmark.
\newblock In {\em Proceedings of the IEEE/CVF conference on computer vision and
  pattern recognition}, pages 10863--10872, 2019.

\bibitem{li2019multi}
Miaopeng Li, Zimeng Zhou, and Xinguo Liu.
\newblock Multi-person pose estimation using bounding box constraint and lstm.
\newblock {\em IEEE Transactions on Multimedia}, 21(10):2653--2663, 2019.

\bibitem{li2021tokenpose}
Yanjie Li, Shoukui Zhang, Zhicheng Wang, Sen Yang, Wankou Yang, Shu-Tao Xia,
  and Erjin Zhou.
\newblock Tokenpose: Learning keypoint tokens for human pose estimation.
\newblock In {\em Proceedings of the IEEE/CVF International Conference on
  Computer Vision}, pages 11313--11322, 2021.

\bibitem{lin2014microsoft}
Tsung-Yi Lin, Michael Maire, Serge Belongie, James Hays, Pietro Perona, Deva
  Ramanan, Piotr Doll{\'a}r, and C~Lawrence Zitnick.
\newblock Microsoft coco: Common objects in context.
\newblock In {\em European conference on computer vision}, pages 740--755.
  Springer, 2014.

\bibitem{liu2021Swin}
Ze Liu, Yutong Lin, Yue Cao, Han Hu, Yixuan Wei, Zheng Zhang, Stephen Lin, and
  Baining Guo.
\newblock Swin transformer: Hierarchical vision transformer using shifted
  windows.
\newblock {\em arXiv preprint arXiv:2103.14030}, 2021.

\bibitem{mao2021tfpose}
Weian Mao, Yongtao Ge, Chunhua Shen, Zhi Tian, Xinlong Wang, and Zhibin Wang.
\newblock Tfpose: Direct human pose estimation with transformers.
\newblock {\em arXiv preprint arXiv:2103.15320}, 2021.

\bibitem{mathis2020deep}
Mackenzie~Weygandt Mathis and Alexander Mathis.
\newblock Deep learning tools for the measurement of animal behavior in
  neuroscience.
\newblock {\em Current opinion in neurobiology}, 60:1--11, 2020.

\bibitem{moon2019posefix}
Gyeongsik Moon, Ju~Yong Chang, and Kyoung~Mu Lee.
\newblock Posefix: Model-agnostic general human pose refinement network.
\newblock In {\em Proceedings of the IEEE/CVF Conference on Computer Vision and
  Pattern Recognition}, pages 7773--7781, 2019.

\bibitem{newell2017associative}
Alejandro Newell, Zhiao Huang, and Jia Deng.
\newblock Associative embedding: End-to-end learning for joint detection and
  grouping.
\newblock {\em Advances in neural information processing systems}, 30, 2017.

\bibitem{qiu2020peeking}
Lingteng Qiu, Xuanye Zhang, Yanran Li, Guanbin Li, Xiaojun Wu, Zixiang Xiong,
  Xiaoguang Han, and Shuguang Cui.
\newblock Peeking into occluded joints: A novel framework for crowd pose
  estimation.
\newblock In {\em European Conference on Computer Vision}, pages 488--504.
  Springer, 2020.

\bibitem{redmon2018yolov3}
Joseph Redmon and Ali Farhadi.
\newblock Yolov3: An incremental improvement.
\newblock {\em arXiv preprint arXiv:1804.02767}, 2018.

\bibitem{ren2015faster}
Shaoqing Ren, Kaiming He, Ross Girshick, and Jian Sun.
\newblock Faster r-cnn: Towards real-time object detection with region proposal
  networks.
\newblock {\em Advances in neural information processing systems}, 28, 2015.

\bibitem{ruggero2017benchmarking}
Matteo Ruggero~Ronchi and Pietro Perona.
\newblock Benchmarking and error diagnosis in multi-instance pose estimation.
\newblock In {\em Proceedings of the IEEE international conference on computer
  vision}, pages 369--378, 2017.

\bibitem{shi2022end}
Dahu Shi, Xing Wei, Liangqi Li, Ye Ren, and Wenming Tan.
\newblock End-to-end multi-person pose estimation with transformers.
\newblock In {\em Proceedings of the IEEE/CVF Conference on Computer Vision and
  Pattern Recognition}, pages 11069--11078, 2022.

\bibitem{tang2018deeply}
Wei Tang, Pei Yu, and Ying Wu.
\newblock Deeply learned compositional models for human pose estimation.
\newblock In {\em ECCV}, pages 190--206, 2018.

\bibitem{wang2022contextual}
Dongkai Wang and Shiliang Zhang.
\newblock Contextual instance decoupling for robust multi-person pose
  estimation.
\newblock In {\em Proceedings of the IEEE/CVF Conference on Computer Vision and
  Pattern Recognition}, pages 11060--11068, 2022.

\bibitem{wang2020deep}
Jingdong Wang, Ke Sun, Tianheng Cheng, Borui Jiang, Chaorui Deng, Yang Zhao,
  Dong Liu, Yadong Mu, Mingkui Tan, Xinggang Wang, Wenyu Liu, and Bin Xiao.
\newblock Deep high-resolution representation learning for visual recognition,
  2020.

\bibitem{xu2022vitpose}
Yufei Xu, Jing Zhang, Qiming Zhang, and Dacheng Tao.
\newblock Vi{TP}ose: Simple vision transformer baselines for human pose
  estimation.
\newblock In Alice~H. Oh, Alekh Agarwal, Danielle Belgrave, and Kyunghyun Cho,
  editors, {\em Advances in Neural Information Processing Systems}, 2022.

\bibitem{xue2022learning}
Nan Xue, Tianfu Wu, Gui-Song Xia, and Liangpei Zhang.
\newblock Learning local-global contextual adaptation for multi-person pose
  estimation.
\newblock In {\em Proceedings of the IEEE/CVF Conference on Computer Vision and
  Pattern Recognition}, pages 13065--13074, 2022.

\bibitem{yang2021transpose}
Sen Yang, Zhibin Quan, Mu Nie, and Wankou Yang.
\newblock Transpose: Keypoint localization via transformer.
\newblock In {\em Proceedings of the IEEE/CVF International Conference on
  Computer Vision}, pages 11802--11812, 2021.

\bibitem{zhang2019pose2seg}
Song-Hai Zhang, Ruilong Li, Xin Dong, Paul Rosin, Zixi Cai, Xi Han, Dingcheng
  Yang, Haozhi Huang, and Shi-Min Hu.
\newblock Pose2seg: Detection free human instance segmentation.
\newblock In {\em Proceedings of the IEEE/CVF Conference on Computer Vision and
  Pattern Recognition}, pages 889--898, 2019.

\end{thebibliography}

\end{document}